\documentclass[10pt,twocolumn,letterpaper]{article}

\usepackage[pagenumbers]{iccv} 
\usepackage{wrapfig}

\usepackage{amsmath,amsfonts,bm}

\usepackage{scalerel}

\def\eqref#1{equation~\ref{#1}}

\def\1{\bm{1}}

\DeclareMathAlphabet{\mathsfit}{\encodingdefault}{\sfdefault}{m}{sl}
\SetMathAlphabet{\mathsfit}{bold}{\encodingdefault}{\sfdefault}{bx}{n}

\DeclareMathOperator*{\argmin}{arg\,min}
\newcommand{\leqnomode}{\tagsleft@true\let\veqno\@@leqno}%
\newcommand{\reqnomode}{\tagsleft@false\let\veqno\@@eqno}%
\newcommand*{\compress}{\@minipagetrue}

\usepackage{graphicx}
\usepackage{subcaption}
\usepackage{url}
\usepackage{textcomp,gensymb}
\usepackage[normalem]{ulem} 

\definecolor{Cerulean}{rgb}{0,0,0.95}
\definecolor{LimeGreen}{rgb}{0.15,0.65,0.15}
\definecolor{RoyalBlue}{rgb}{0.25,0.41,0.88}
\definecolor{Rose}{rgb}{1.0, 0.15, 0.21}
\definecolor{Orange}{rgb}{1.0, 0.5, 0.0}
\definecolor{Gray}{gray}{0.6}
\definecolor{Black}{gray}{0.0}
\definecolor{Purple}{rgb}{0.77,0.12,0.64}
\definecolor{FullBlue}{rgb}{0,0,1}
\usepackage{algorithm}
\usepackage{algpseudocode}
\usepackage{amsmath}
\usepackage{graphicx} 
\usepackage{algpseudocode}
\usepackage{booktabs}
\usepackage{multicol}
\usepackage{multirow}
\usepackage{colortbl}
\usepackage{placeins}    
\colorlet{mygray}{black!10}
\usepackage{float}
\definecolor{TableBlue}{rgb}{0.17,0.49,0.75}
\definecolor{Cerulean}{rgb}{0,0,0.95}
\definecolor{LimeGreen}{rgb}{0.15,0.65,0.15}
\definecolor{RoyalBlue}{rgb}{0.25,0.41,0.88}
\definecolor{Rose}{rgb}{1.0, 0.15, 0.21}
\definecolor{Orange}{rgb}{1.0, 0.5, 0.0}
\definecolor{Gray}{gray}{0.6}
\definecolor{Black}{gray}{0.0}
\definecolor{Purple}{rgb}{0.77,0.12,0.64}

\definecolor{iccvblue}{rgb}{0.21,0.49,0.74}
\usepackage[pagebackref,breaklinks,colorlinks,allcolors=iccvblue]{hyperref}

\title{AutoScale: 
Linear Scalarization Guided by Multi-Task Optimization Metrics
}

\author{
    Yi Yang$^{1,2}$\thanks{Equal contribution. {\tt\scriptsize{\{yiya,ikemura\}@kth.se}}} \quad Kei Ikemura$^{1}$\footnotemark[1] \quad Qingwen Zhang$^{1}$ \quad Xiaomeng Zhu$^{1,2}$ \quad Ci Li$^{1}$ \\
    Nazre Batool$^{3}$ \quad Sina Sharif Mansouri$^{2}$ \quad John Folkesson$^{1}$ 
    \\
KTH Royal Institute of Technology, Sweden$^{1}$ \\\quad Scania AB, Sweden$^{2}$ \quad National University of Ireland, Galway$^{3}$ \quad \\
}

\begin{document}
\maketitle
\begin{abstract}
Recent multi-task learning studies suggest that linear scalarization, when using well-chosen fixed task weights, can achieve comparable to or even better performance than complex multi-task optimization (MTO) methods.
It remains unclear why certain weights yield optimal performance and how to determine these weights without relying on exhaustive hyperparameter search.
This paper establishes a direct connection between linear scalarization and MTO methods, revealing through extensive experiments that well-performing scalarization weights exhibit specific trends in key MTO metrics, such as high gradient magnitude similarity.
Building on this insight, we introduce AutoScale, a simple yet effective two-phase framework that uses these MTO metrics to guide weight selection for linear scalarization, without expensive weight search.
AutoScale consistently shows superior performance with high efficiency across diverse datasets including a new large-scale benchmark.
\end{abstract}

\section{Introduction}
\label{sec:intro}

Multi-task learning (MTL) has gained significant attention, due to its efficiency in using a single network to learn multiple tasks, and potential knowledge transfer among tasks \citep{GradNorm, FAMO,xin2022current,hu2024revisiting}.
However, MTL suffers from task conflicts, such as varying task difficulties, imbalanced gradient scale, and conflicting optimization directions.
To address these challenges, researchers have developed Multi-Task Optimization (MTO) methods that dynamically adjust task gradients and/or losses to balance learning~\citep{GradNorm, IMTL, PCGrad, AlignedMTL,ban2024fair,lin2024smooth, NEURIPS2021_b2397517,liu2022auto,xiao2023direction}.

Interestingly, recent studies~\citep{xin2022current, kurin2022defense, royer2024scalarization} have shown that simple \textit{linear scalarization}, which minimizes a weighted sum of task losses with fixed weight coefficients, can perform comparably to or even surpass complex MTO methods - when the weights are carefully chosen. 
Despite the conceptual and operational simplicity of linear scalarization, 
its performance relies on selecting optimal weights, requiring a costly hyperparameter search.
This raises questions about our understanding of linear scalarization, particularly: 1) why does it perform well under certain weights and, 2) how to determine these weights efficiently. 

Toward answering these questions, we propose a novel perspective: the effectiveness of linear scalarization is closely linked to the metrics used in MTO methods, which were originally developed independently. Prior MTO research has developed various heuristics to quantify optimization challenges and ensure balanced training, including gradient magnitude balance~\citep{GradNorm, MGDA, IMTL}, gradient direction alignment~\citep{PCGrad, GradNorm, MGDA, IMTL}, condition number~\citep{AlignedMTL}, etc.
In this work, we systematically examine these heuristics together, referring to them as \textit{MTO metrics}. Through extensive empirical analysis, we find that well-performing linear scalarization consistently aligns with certain MTO metrics, as depicted in \Cref{fig:cost_city}, while other metrics show weak or inconsistent correlations. This suggests that optimal scalarization weights naturally preserve certain beneficial MTO metric properties, offering a systematic way to guide weight selection. 

\begin{figure*}[t]
\begin{subfigure}{0.245\linewidth}
    \centering
    \includegraphics[width=\linewidth]{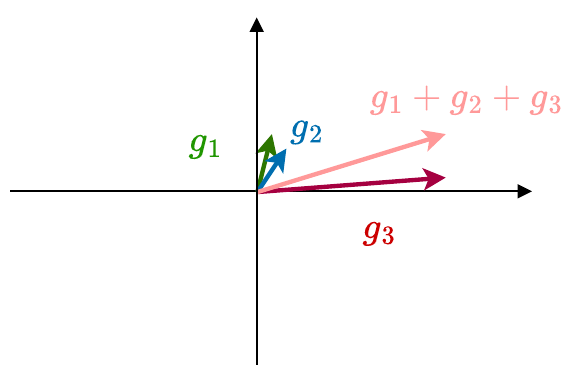}
    \caption{\scriptsize{Gradient dominance}}
    \label{fig:mtl-issue-grad-dom}
\end{subfigure}
\begin{subfigure}{0.245\linewidth}
    \centering
    \includegraphics[width=\linewidth]{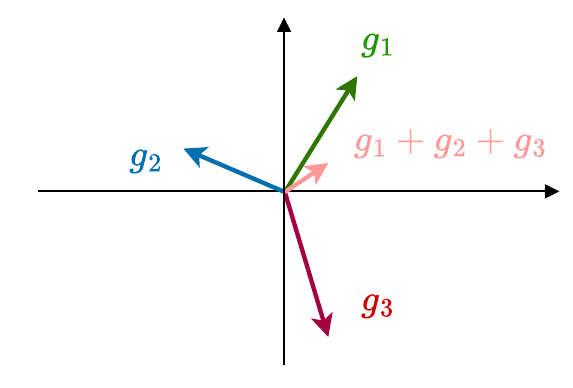}
    \caption{\scriptsize{Gradient conflict}}
    \label{fig:mtl-issue-grad-conf}
\end{subfigure}
\begin{subfigure}{0.245\linewidth}
    \centering
    \includegraphics[width=\linewidth]{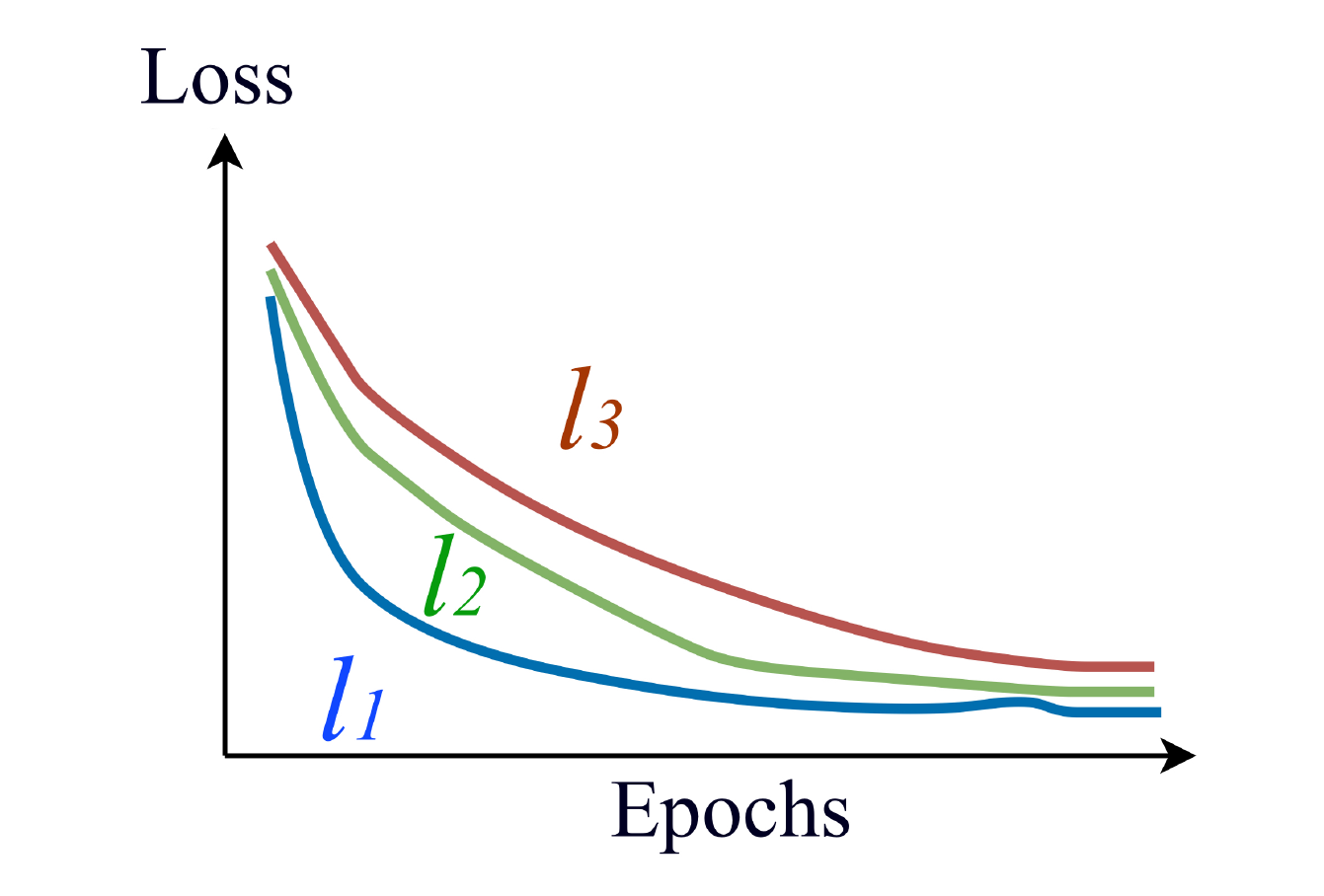}
    \caption{\scriptsize{Imbalanced convergence}}
    \label{fig:mtl-issue-convg-spd}
\end{subfigure}
\begin{subfigure}{0.245\linewidth}
    \centering
    \includegraphics[width=\linewidth]{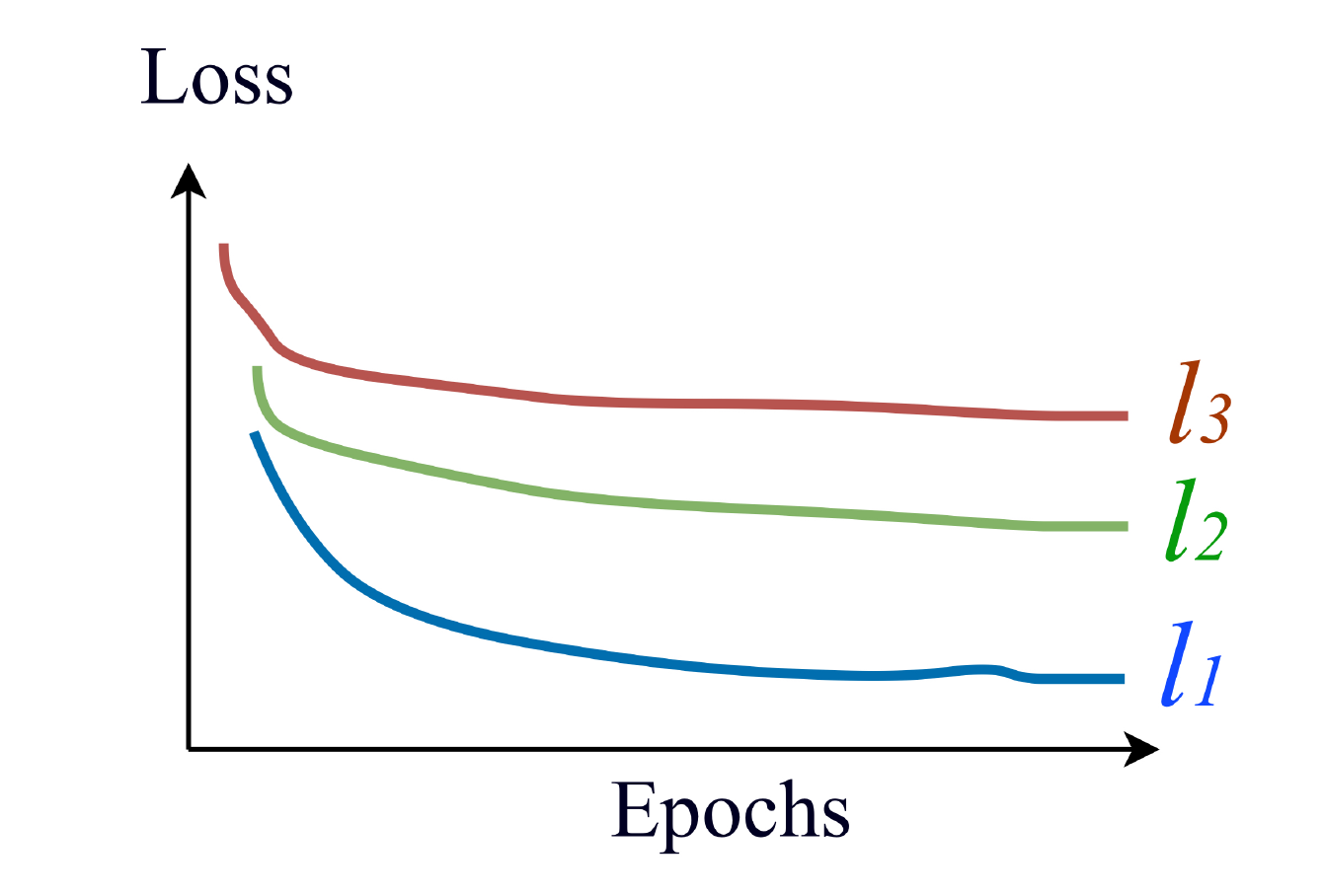}
    \caption{\scriptsize{Imbalanced loss}}
    \label{fig:mtl-issue-loss-imb}
\end{subfigure}
\caption{Illustration showing different multi-task training issues, $g_k$ and $l_k$ represent gradient and loss for task $k$ respectively.}
\vspace{-10pt}
\end{figure*}
Based on this insight, we propose AutoScale, a two-phase framework that eliminates costly hyperparameter searches by leveraging aligned MTO metrics for automatic weight selection. In the first phase, AutoScale optimizes a selected MTO metric using gradient and loss information during training. In the second phase, it applies the identified weights for scalarization, ensuring efficiency.
Notably, AutoScale is highly adaptable, as it can integrate any MTO metric to guide weight selection.

We evaluate AutoScale on three aligned MTO metrics: gradient magnitude similarity, condition number, and loss scale balance—as they capture distinct training challenges in multi-task learning.
We demonstrate its effectiveness across diverse datasets, including the widely used MTL benchmarks CityScapes~\citep{cordts2016cityscapes} and NYUv2~\citep{silberman2012indoor}, as well as a large-scale autonomous driving dataset NuScenes.
We evaluate state-of-the-art MTO methods on NuScenes, introducing a new large-scale MTL benchmark.

To the best of our knowledge, this work is the first to establish a direct connection between linear scalarization and MTO metrics. Our key contributions are as follows:

\begin{itemize}
    \item We show that the effectiveness of linear scalarization weights is strongly correlated with certain MTO metrics, such as high gradient similarity among tasks and low condition number, providing a principled perspective on weight selection.

    \item We propose AutoScale, a two-phase framework that optimizes MTO metrics to automatically determine scalarization weights, without the need for costly weight search.

    \item We demonstrate that AutoScale consistently achieves superior performance and high efficiency compared to SOTA methods, across various datasets including a new large-scale autonomous driving MTL benchmark dataset.
    
\end{itemize}

\vspace{3pt}
 
Our code will be publicly available upon publication.
\vspace{-1pt}
\section{Related Work}
\subsection{Multi-Task Learning: Overview}

Research in multi-task learning, particularly within deep learning, has largely focused on three main directions: (1) MTL-specific architectures, (2) task grouping, and (3) Multi-Task Optimization algorithms (MTOs).
MTL-specific architecture aims to improve performance by designing network structures to better handle multiple tasks~\citep{Misra_2016_CVPR, dai2016instance, long2017learning, taskprompter2023}.
Task grouping, on the other hand, explores the relationships among tasks and reduces negative transfer by grouping non-conflicting or minimally conflicting tasks during training~\citep{thrun1996discovering, zamir2018taskonomy, WhichTasks, shen2024go4aligngroupoptimizationmultitask,jeong2025selectivetaskgroupupdates}. 
Lastly, MTOs address the problem by dynamically manipulating gradients and loss to update network parameters~\citep{GradNorm, AlignedMTL,FAMO, quinton2024jacobian, liu2024online, achituve2024bayesianuncertaintygradientaggregation, lin2023dualbalancingmultitasklearning, fernando2023mitigating, chen2023three}. 
In this work, we focus on the last paradigm.

\subsection{Multi-Task Optimization} 
\label{sec:mto}
Over the past several years, numerous MTO methods have been proposed where the majority is built upon different challenges in MTL training. Here, we categorize these training challenges into five types and provide related literature for each:
(1) gradient dominance, (2) gradient conflict, (3) imbalanced convergence speed, (4) imbalanced loss, and (5) instability. 

\paragraph{Gradient Dominance.} Variations in the scale of task-wise gradients on the shared parameters create impartial learning outcomes~\citep{IMTL}, where the network converges primarily on tasks with higher gradient magnitudes (as shown in \Cref{fig:mtl-issue-grad-dom}). 
To address this, GradNorm~\citep{GradNorm} dynamically adjusts task weights to ensure that the norm of each task's scaled gradient is balanced.
On the other hand, IMTL-G~\citep{IMTL} tackles this problem by finding an aggregated gradient with equal projections onto each task gradient.
\vspace{-11pt}
\paragraph{Gradient Conflict.} Conflicting gradients with opposing directions (as shown in \Cref{fig:mtl-issue-grad-conf}) can cause negative transfers~\citep{AlignedMTL, lee2018deep}. CosReg~\citep{CosReg} proposes a regularization term based on the squared cosine similarity between task gradients, penalizing the network when conflicting gradients are generated. PCGrad~\citep{PCGrad}, instead, avoids task conflicts by projecting the gradient of one task onto the normal plane of another. Similarly, \citet{CAGrad} find a conflict-averse direction to minimize overall conflict, while GradDrop~\citep{GradDrop} enforces the sign consistency across task gradients.
\citet{NashMTL} solve it as a Nash bargaining game.

\vspace{-11pt}
\paragraph{Imbalanced Convergence Speed.} 
Different tasks have varying levels of difficulty, potentially leading to different convergence speeds~\citep{DTP,AMTL} as shown in \Cref{fig:mtl-issue-convg-spd}.
To address this issue, methods like GradNorm~\citep{GradNorm}, DTP~\citep{DTP}, DWA~\citep{DWA}, AMTL~\citep{AMTL} and ExcessMTL~\citep{he2024robust} define specific indicators of training convergence and adjust task weights based on these indicators.
Additionally, \citet{AFD-OTW} propose to train single-task networks alongside the MTL network, and use the convergence speed of the single-task network to guide online knowledge distillation.

\begin{figure*}[t!]
\centering

\begin{subfigure}[t]{1\textwidth}
\centering
    \includegraphics[width=0.8\textwidth]{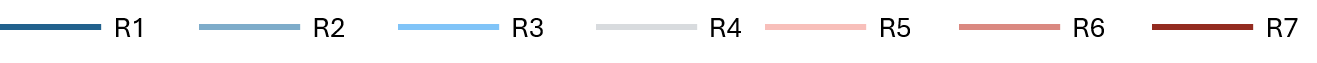}
\end{subfigure}
\begin{subfigure}[t]{.32\linewidth}
    \centering
    \includegraphics[width=0.93\linewidth]{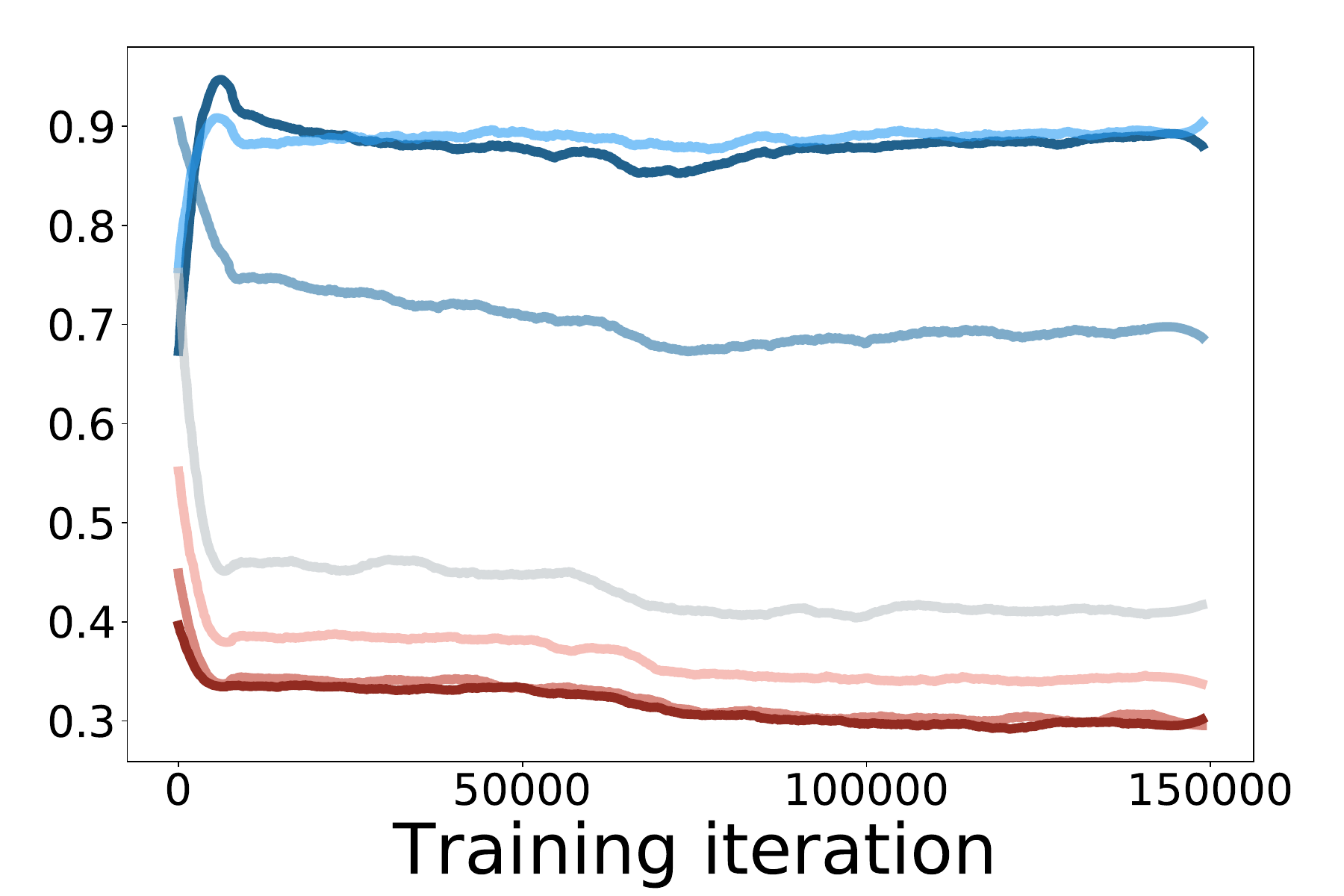}
    \caption{Gradient magnitude similarity $(M_{\text{GMS}}^t)$}
    \label{fig:city:one:gms}
\end{subfigure}
\hfill
\begin{subfigure}[t]{.32\linewidth}
\centering
    \includegraphics[width=0.93\linewidth]{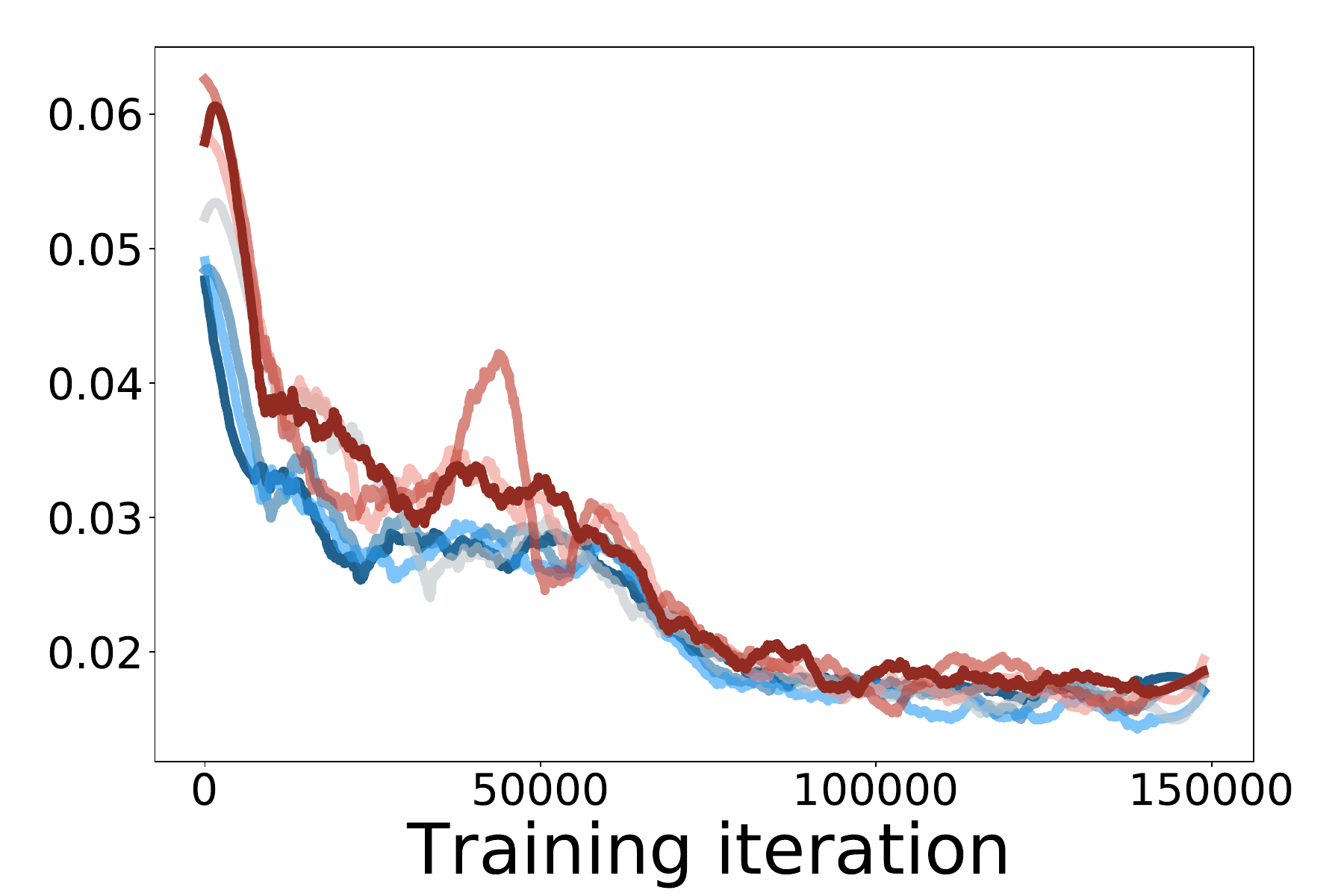}
    \caption{Gradient cosine similarity $(M_{\text{GCS}}^t)$}
    \label{fig:city:2:cosdist}
\end{subfigure}
\hfill
\begin{subfigure}[t]{.32\linewidth}
    \centering
    \includegraphics[width=0.93\linewidth]{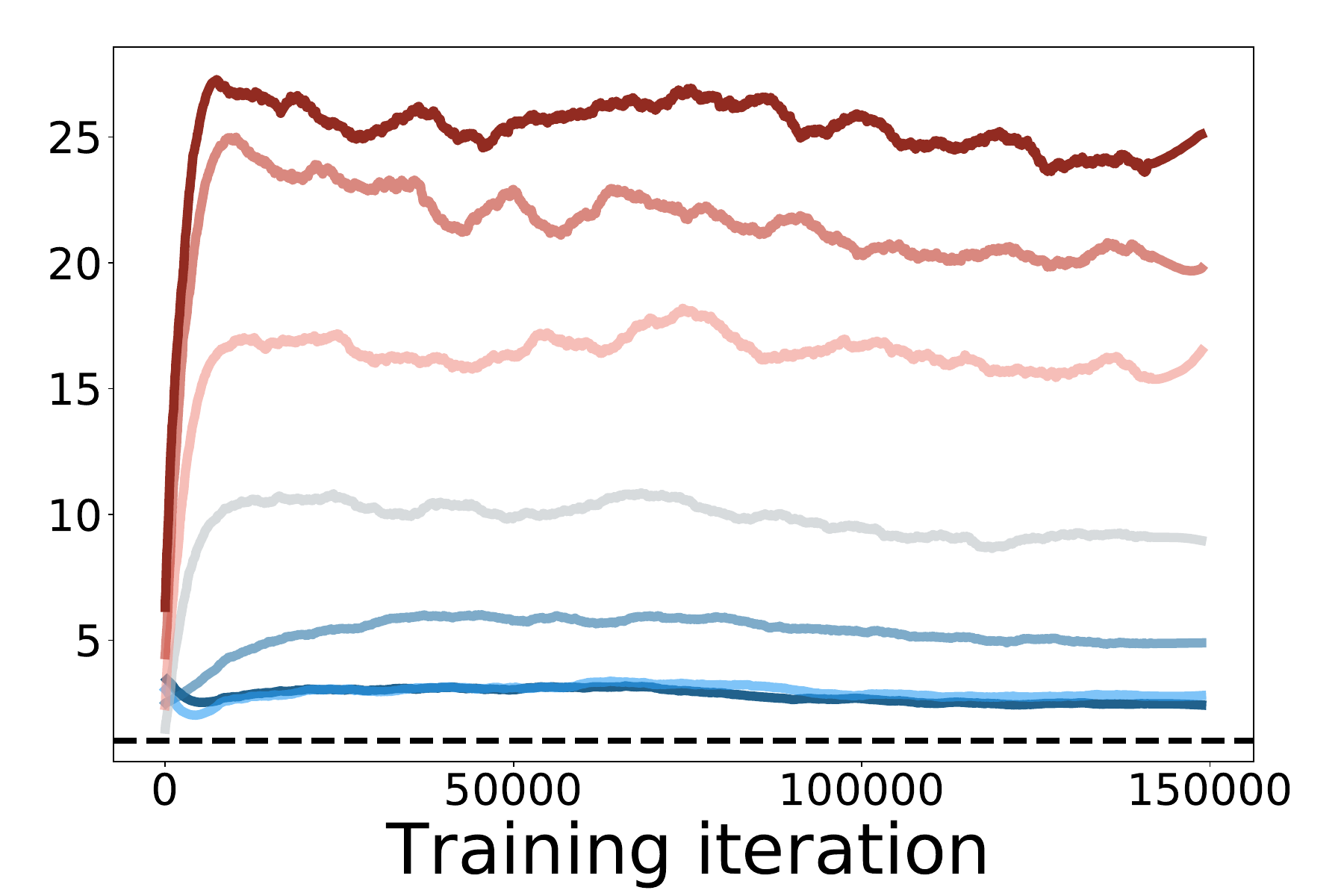}
    \caption{Condition number $(M_{\text{CN}}^t)$}
    \label{fig:city:3:cnumber}
\end{subfigure}

\begin{subfigure}[t]{.32\linewidth}
    \centering
    \includegraphics[width=0.93\linewidth]{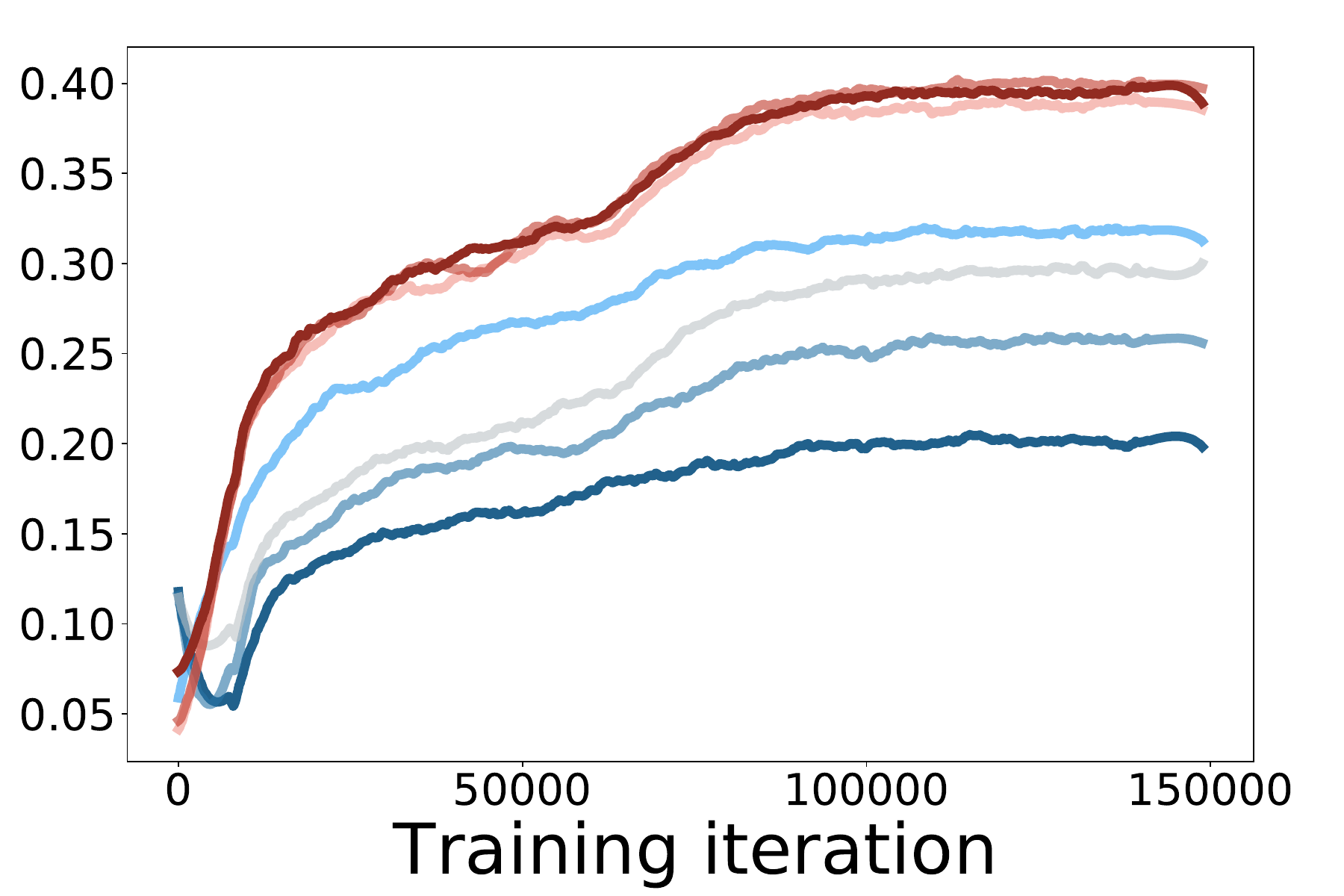}
    \caption{Std of inverse learning rate  $(M_{\text{ILR}}^t)$}
    \label{fig:city:4:ilr}
\end{subfigure}
\hfill
\begin{subfigure}[t]{.32\linewidth}
    \centering
    \includegraphics[width=0.93\linewidth]{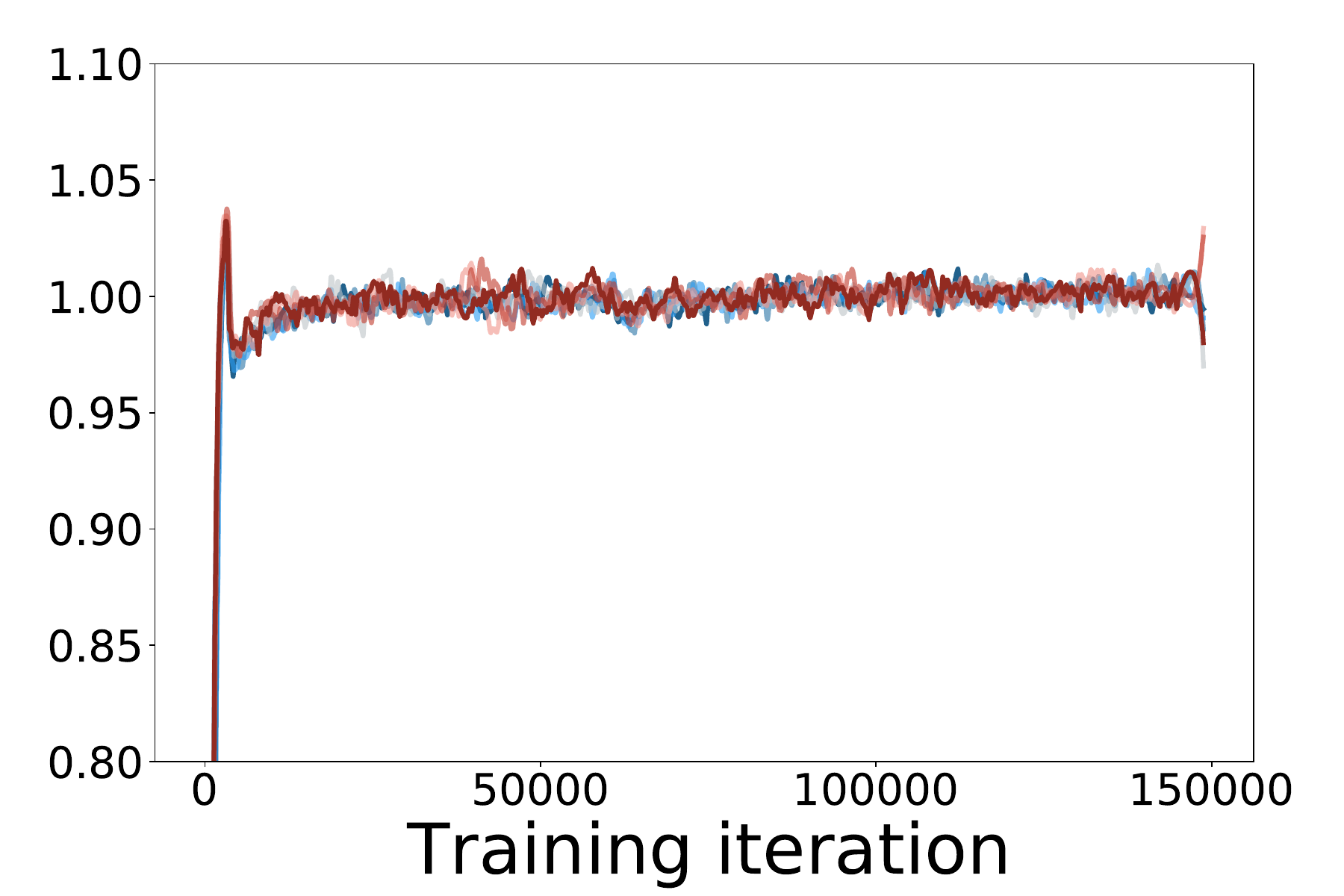}
    \caption{Descending rate $(M_{\text{LDR}}^t)$}
    \label{fig:city:5:ldr}
\end{subfigure}
\hfill
\begin{subfigure}[t]{.32\linewidth}
\centering
    \includegraphics[width=0.93\linewidth]{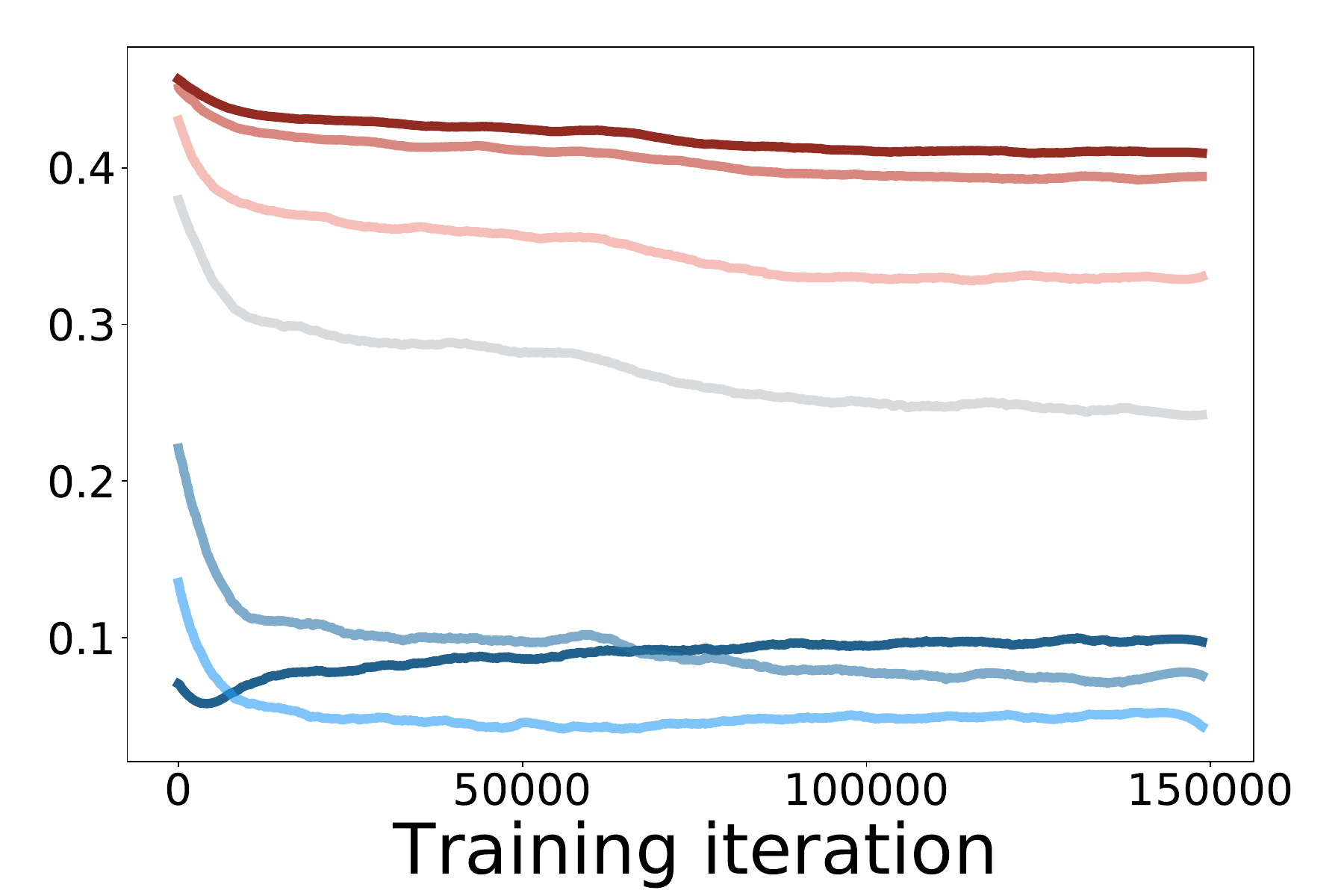}
    \caption{Std of relative loss $(M_{\text{RL}}^t)$}
    \label{fig:city:6:rloss}
\end{subfigure}
\caption{Evaluation on different MTO metrics and how they evolve during the training process of seven linear scalarization weight sets on the CityScapes dataset, with the performance ranking: R1 $>$ R2 $>$ R3 $>$ R4 $>$ R5 $>$ R6 $>$ R7. The metrics include (a) gradient dominance: gradient magnitude similarity; (b) gradient conflict: gradient cosine similarity; (c) instability: condition number; (d,e) imbalance convergence speed: inverse learning rate, loss descending rate; (f) imbalanced loss: relative loss scale.
The metrics are defined in \Cref{sec:mtl-met-overview}.
\scriptsize{$\star$Unless specified, the metric values represent the average across tasks (or task pairs for metrics like similarity); captions with ``std''- indicate the standard deviation across tasks. The performance is ranked by $\mathbf{\Delta m}$: measuring the average performance drop across tasks, as detailed in \Cref{sec:eval_metrics}.}}
\label{fig:eval_city}
\vspace{-10pt}
\end{figure*}
\vspace{-11pt}
\paragraph{Imbalanced Loss.} Imbalances in the scale of task-specific losses (shown in \Cref{fig:mtl-issue-loss-imb}) can result in suboptimal training outcomes. Many works focus on equalizing the scale of task losses. GLS~\citep{GLS} adopts geometric mean to prevent tasks with larger losses from dominating the overall loss function. 
Following GLS, \citet{AMTL} propose a weighted geometric mean of loss that is robust to scale variation.
\citet{IMTL} propose IMTL-L to re-scale and balance task losses by deriving tasks weights. 

\vspace{-11pt}
\paragraph{Instability.} Aligned-MTL~\citep{AlignedMTL} defines the instability in MTL training as the instability of the linear system composed by task gradients. It proposes to stabilize the training process by aligning the principal components of the gradient matrix.

We further provide formal definitions of the MTO metrics introduced by existing MTO methods to quantify and analyze these five MTL issues in~\Cref{sec:mtl-met-overview}.

\vspace{-2pt}
\subsection{Revisiting Linear Scalarization}
In recent years, linear scalarization has been revisited and argued to be an appealing alternative to more complex MTOs.
Although linear scalarization has been shown to fail to identify solutions that lie on the non-convex part of the Pareto front \citep{hu2024revisiting}, many studies~\citep{kurin2022defense, xin2022current, elich2024challenging, royer2024scalarization} demonstrate that, in practice, it achieves performance comparable to or even better than other MTOs through large-scale experiments.
However, a major open challenge for linear scalarization is identifying the optimal set of scalarization weights with minimal computational overhead.
Despite the fact that more efficient search methods have been proposed~\citep{royer2024scalarization}, they remain costly
compared to directly applying existing MTOs due to requiring multiple training runs.
In this work, we address the problem by proposing a unified pipeline that efficiently localizes optimal scalarization weights within \textit{a single training run}.

\section{Notation}
In this section, we provide the formal definition of MTL and notations used in this work.

The objective of MTL is to optimize $K>1$ tasks simultaneously using a single network with parameters $\theta$,
where each task $k$ has a corresponding loss function $l_k(\theta)$. 

We refer to \textit{linear scalarization weight} as a weight vector
$\mathrm{w}=[w_1 \; w_2 \; ... \; w_K] \in \mathbb{R}_+^K$ where, at each training iteration $t$, the total loss is given as:
\vspace{-10pt}
\begin{equation}
    l^t = \sum_{k=1}^K w_k l_k^t,
    \vspace{-10pt}
\end{equation}
\noindent where $l_k^t$ denotes the loss function value of task $k$ at time $t$. The dependency on $\theta$ is omitted for simplicity.

Next, let us define $g_k^t = \nabla_{\theta_{\text{shared}}} l_k^t$ to be the gradient for task $k$, w.r.t. shared parameters $\theta_{\text{shared}}$ at time $t$; $\mathbf{G}^t=[g_1^t \; g_2^t \; ... \; g_K^t]$ to be gradient matrix at time $t$; and $\mathbf{L}^t = [l_1^t \; l_2^t \; ... \; l_K^t]$ to be the vector of task losses at time $t$.

\begin{figure*}[t!]
\centering
\begin{subfigure}[t]{.32\linewidth}
    \centering
    \includegraphics[width=1.0\linewidth]{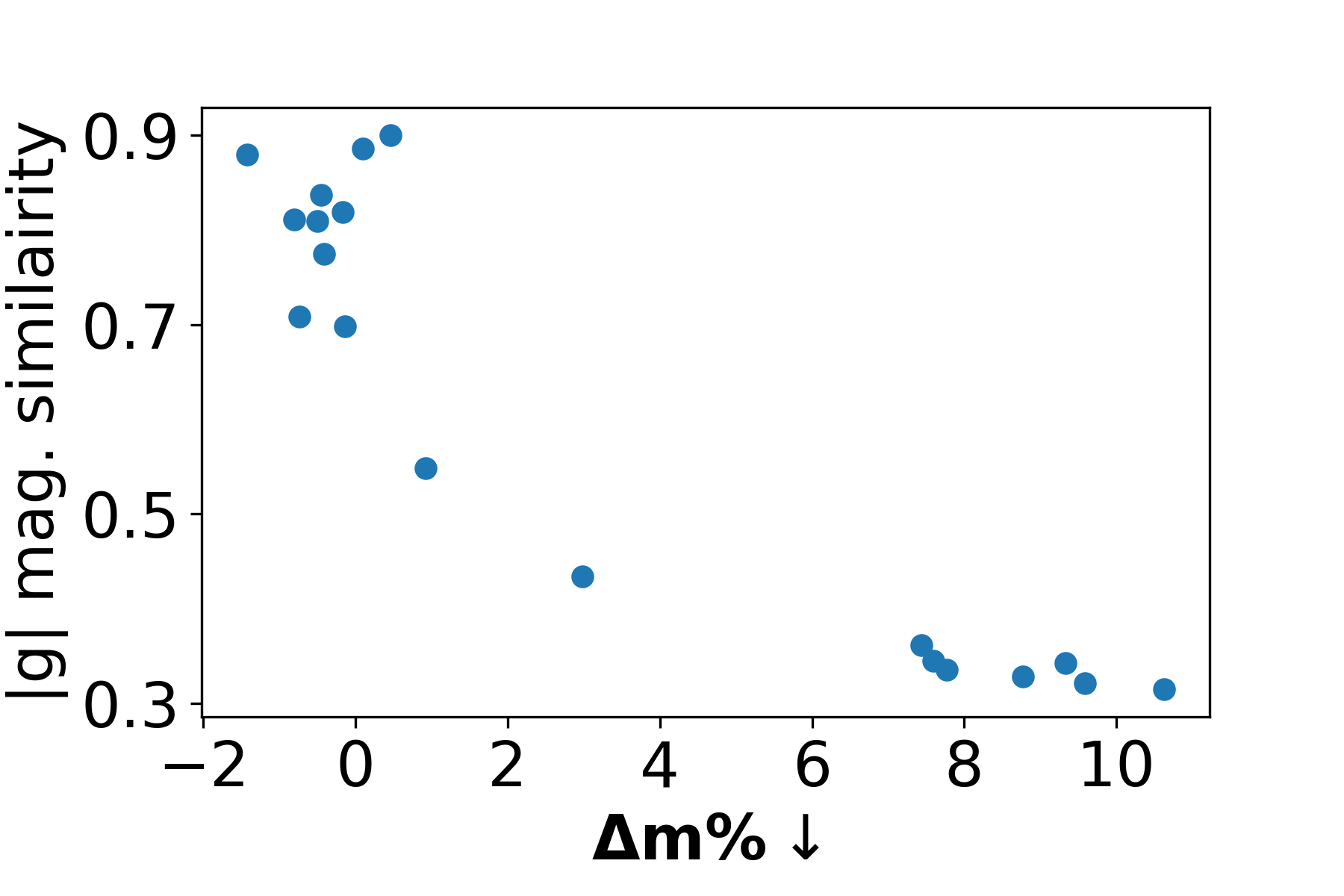}
    \caption{Gradient magnitude similarity}
    \label{fig:city:cost:gms}
\end{subfigure}
\hfill
\begin{subfigure}[t]{.32\linewidth}
\centering
    \includegraphics[width=1.0\linewidth]{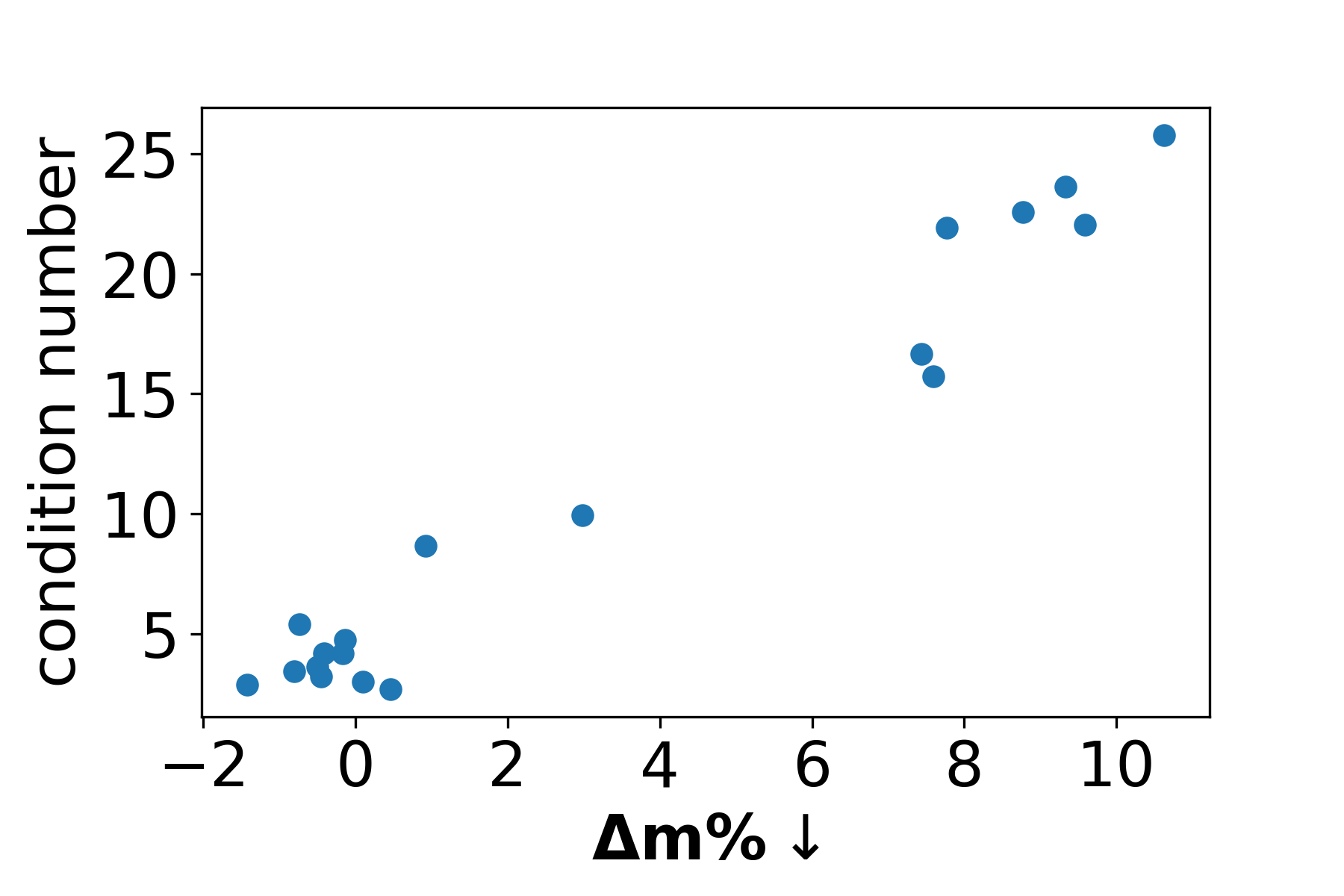}
    \caption{Condition number}
    \label{fig:city:cost:cnumber}
\end{subfigure}
\hfill
\begin{subfigure}[t]{.32\linewidth}
    \centering
    \includegraphics[width=1.0\linewidth]{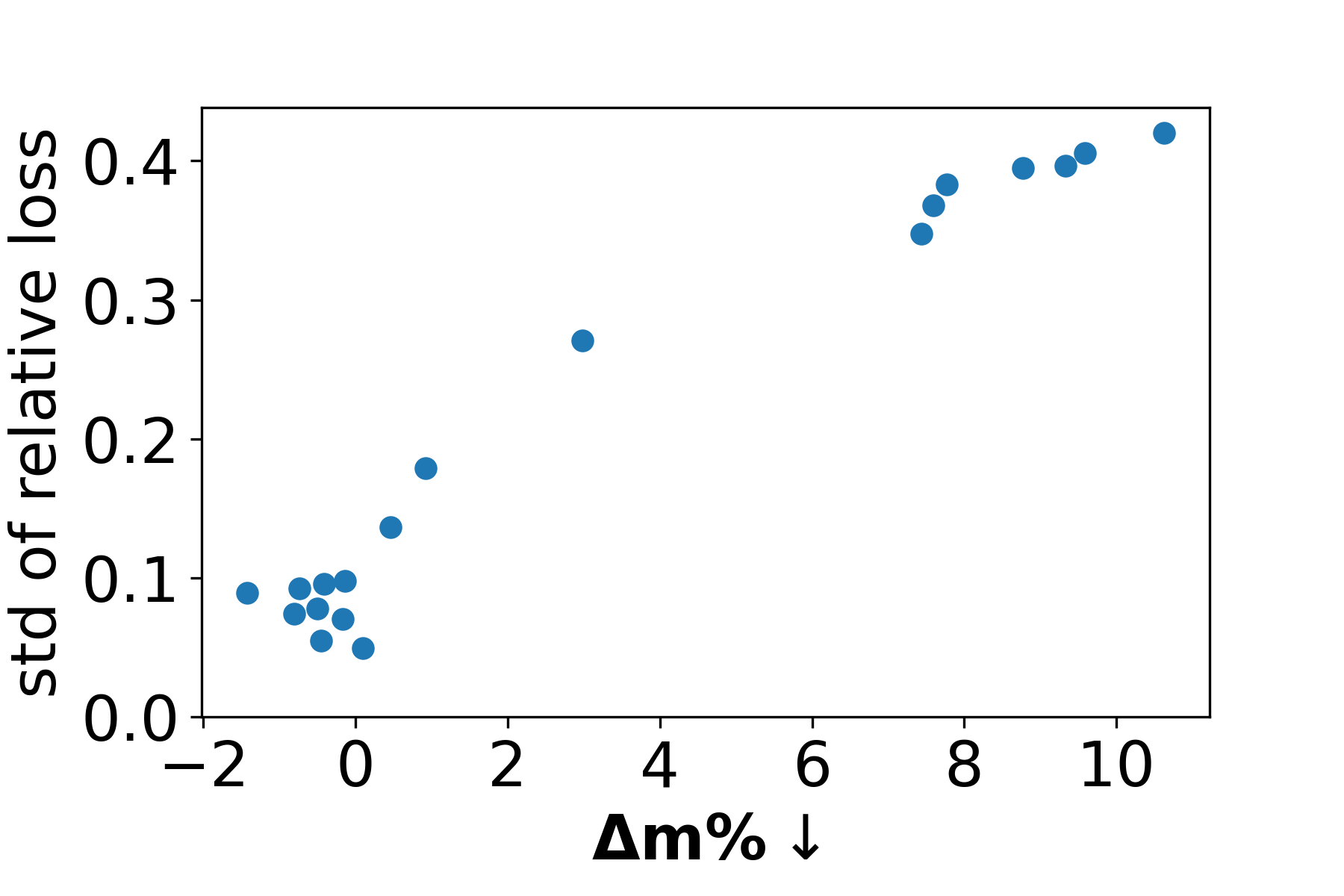}
    \caption{Std of relative loss}
    \label{fig:city:cost:std_loss}
\end{subfigure}
\caption{Performance ($\mathbf{\Delta m}\downarrow$: average of performance drop compared to single-task learning, lower value indicate higher performance, as detailed in \Cref{sec:eval_metrics}) vs. metrics values (average over training iterations) of 19 weight sets of linear scalarization.
The plots illustrate a clear correlation between high performance and high gradient magnitude similarity, low condition number, and low standard deviation of relative loss among tasks.
}
\label{fig:cost_city}
\vspace{-10pt}
\end{figure*}

\section{Linear Scalarization Analysis}
\subsection{Analysis Setup}
\label{sec:mtl-met-overview}
To investigate the training dynamics of linear scalarization and understand why certain weights yield strong performance, we conduct large-scale experiments across diverse task weights.
For the first time, we systematically analyze differences between high- and low-performance linear scalarization using a range of MTO metrics introduced in previous MTO methods.

To formalize the analysis, we conduct $N$ independent runs of linear scalarization using a set of random weights $\mathcal{W} = \{\mathrm{w}_1, ..., \mathrm{w}_N\}$, and rank them by performance, where $\mathrm{w}_1$ corresponds to the best performing weight and $\mathrm{w}_N$ to the worst.
During each run, we monitor a total number of \( P \) MTO metrics, represented as \( \mathcal{M} = \{M_1, ..., M_P\} \). The trajectory of each metric \( M_i \) across training steps \( t \) is recorded as \( M^t_i(\mathrm{w})  \).

In this work, we select ten MTO metrics capturing different challenges for evaluation.
We provide formal mathematical definitions for six metrics below, while we summarize and categorize a complete list of metrics
in the supplementary.

\vspace{-11pt} \paragraph{Gradient Dominance.} 
Gradient magnitude similarity quantifies how balanced the gradient magnitudes are between  task pairs~\citep{FULLER, PCGrad} and is defined as:
$$
M_{\text{GMS}}^t(\mathrm{w}_i) =\frac{2 |g_i^t|  \cdot |g_j^t|}{|g_i^t|^2 + |g_j^t|^2} \in [0,1], \quad \forall i \neq j
$$
where a higher value indicates more balanced gradients, while a lower value suggests dominance by one task.

\vspace{-11pt} \paragraph{Gradient Conflict.} 
Gradient conflict is measured via cosine similarity between task gradient directions~\citep{CosReg, AlignedMTL, PCGrad}:
$$
M_{\text{GCS}}^t(\mathrm{w}_i) =\frac{g_i^{t\top} g_j^t}{|g_i^t||g_j^t|} \in [-1,1], \quad \forall i \neq j
$$
where negative values indicate conflicting gradients.

\vspace{-11pt} \paragraph{Imbalanced Convergence Speed.} 
The training progress has been quantified in different ways in research literature. For example, GradNorm~\citep{GradNorm} defines inverse learning rate, $M_{\text{ILR}}$, to be the ratio of current loss to initial loss; while DWA~\citep{DWA} and FAMO~\citep{FAMO} use the ratio of two consecutive steps as the loss descending rate $M_{\text{LDR}}$.
$$
 M_{\text{ILR}}^t(\mathrm{w}_i) = l^t / l^0, ~~~M_{\text{LDR}}^t(\mathrm{w}_i) = l^t / l^{t-1}
$$

\vspace{-11pt} \paragraph{Imbalanced Loss.} 
Loss balance evaluates whether tasks have comparable loss magnitudes.
An intrinsic metric is the relative loss scale:
$$
M_{\text{RL}}^t(\mathrm{w}_i) = \frac{l_i^t}{\sum_{j=1}^K l_j^t} 
$$

\vspace{-11pt} \paragraph{Instability.}
The condition number of gradient matrix, proposed by ~\citep{AlignedMTL} to assess training instability, is given as below:
\[
M_{\text{CN}}^t(\mathrm{w}_i) = \frac{\sigma_{\max}(\mathbf{G}^t)}{\sigma_{\min}(\mathbf{G}^t)}
\]
where $\sigma$ are the singular values of the gradient matrix $\mathbf{G}^t$.
A large condition number indicates unstable training.

\subsection{MTO Metrics Under Scalarization Weights} \label{sec:observed_trend}
The trajectories of various MTO metrics during training in some of our experiments are shown in \Cref{fig:eval_city}.

Interestingly, we find that certain metrics, including gradient magnitude similarity, condition number, and relative loss scale, demonstrate clear patterns that distinguish high-performance from low-performance linear scalarization.
We group these as \textit{key MTO metrics}.
Specifically, better-performing weights exhibit higher gradient magnitude similarity (\Cref{fig:city:one:gms}).
For training instability, the condition number of the best-performing linear scalarization is the lowest, approaching 1 (\Cref{fig:city:3:cnumber}).
Regarding training progress and loss balance, high-performance weights lead to a more balanced convergence and loss scale across tasks, as reflected by smaller standard deviations in inverse learning rates (\Cref{fig:city:4:ilr}) and relative loss scales (\Cref{fig:city:6:rloss}).
\Cref{fig:cost_city} further illustrates the strong correlations between linear scalarization performance and key MTO metrics across 19 runs with different weight sets.

In contrast, some MTO metrics show weak or inconsistent correlation with linear scalarization performance.
For example, the loss descending rate is less informative as depicted in \Cref{fig:city:5:ldr}.
Additionally, since linear scalarization only scales per-task loss, it does not affect the angle (cosine similarity) of the gradients between tasks (\Cref{fig:city:2:cosdist}). 

More MTO metrics visualizations across various datasets are provided in supplementary.

\section{Method: AutoScale}
Given the strong correlation between the performance of linear scalarization and the key MTO metrics, we hypothesize that these metrics serve as reliable indicators for selecting optimal scalarization weights.
Building on this, the weight selection process is formulated as an optimization problem:
\vspace{-6pt}
\begin{equation}
    \mathrm{w}^* = \argmin_\mathrm{w} \mathbf{F}(\mathrm{w}|\mathcal{G}, \mathcal{L}), \quad \textrm{s.t.} \;\,  \sum_{i=1}^K w_i = K, \label{eq:objective}
\end{equation} 

\noindent where the condition variables $\mathcal{G} = \{\mathbf{G}^{t_1}, \mathbf{G}^{t_2}, ..., \mathbf{G}^{t_T}\}$ and $\mathcal{L} = \{\mathbf{L}^{t_1}, \mathbf{L}^{t_2}, ..., \mathbf{L}^{t_T}\}$ represent the set of gradient matrices and task loss vectors sampled over the entire training process, respectively. 
$\mathbf{F}(\mathrm{w})$ is the cost function that evaluates linear scalarization weights $\mathrm{w}$ using selected MTO metrics, computed based on gradients and/or task losses.
It penalizes weights that result in undesirable MTO metric values, such as a high condition number.
By minimizing $\mathbf{F}(\mathrm{w})$, we aim to find the optimal weights that align with favorable MTO metric trends, as discussed in \Cref{sec:observed_trend}.

To solve Equation \ref{eq:objective} and keep efficiency in MTL, we propose AutoScale, an two-stage framework that first approximates the optimal weight $\hat{\mathrm{w}} \approx \mathrm{w}^*$ and then apply it for linear scalarization within \textit{a single training run}.

Algorithm~\ref{algo:autoscale} outlines the procedure of AutoScale. The entire training process is partitioned into two phases based on the exploration ratio $\alpha \in [0, 1]$, which determines the proportion allocated to the exploration phase.
First, in the \textit{exploration phase}, we divide the training iterations into disjoint windows. Within each local window with index $i$, the network is trained with previously estimated weight $\mathrm{w}^{i-1}$; and a new locally optimal weight $\mathrm{w}^i$ is computed by minimizing $\mathbf{F}(\mathrm{w})$ over the window, based on the collected task gradients and losses.
Next, in the \textit{linear scalarization phase}, we estimate an optimal weight $\hat w$ as the average of the last $\eta$ local weights from the exploration phase.
The estimated weight is then applied for the remaining training.

\begin{figure}
    \centering
    \includegraphics[width=0.95\linewidth]{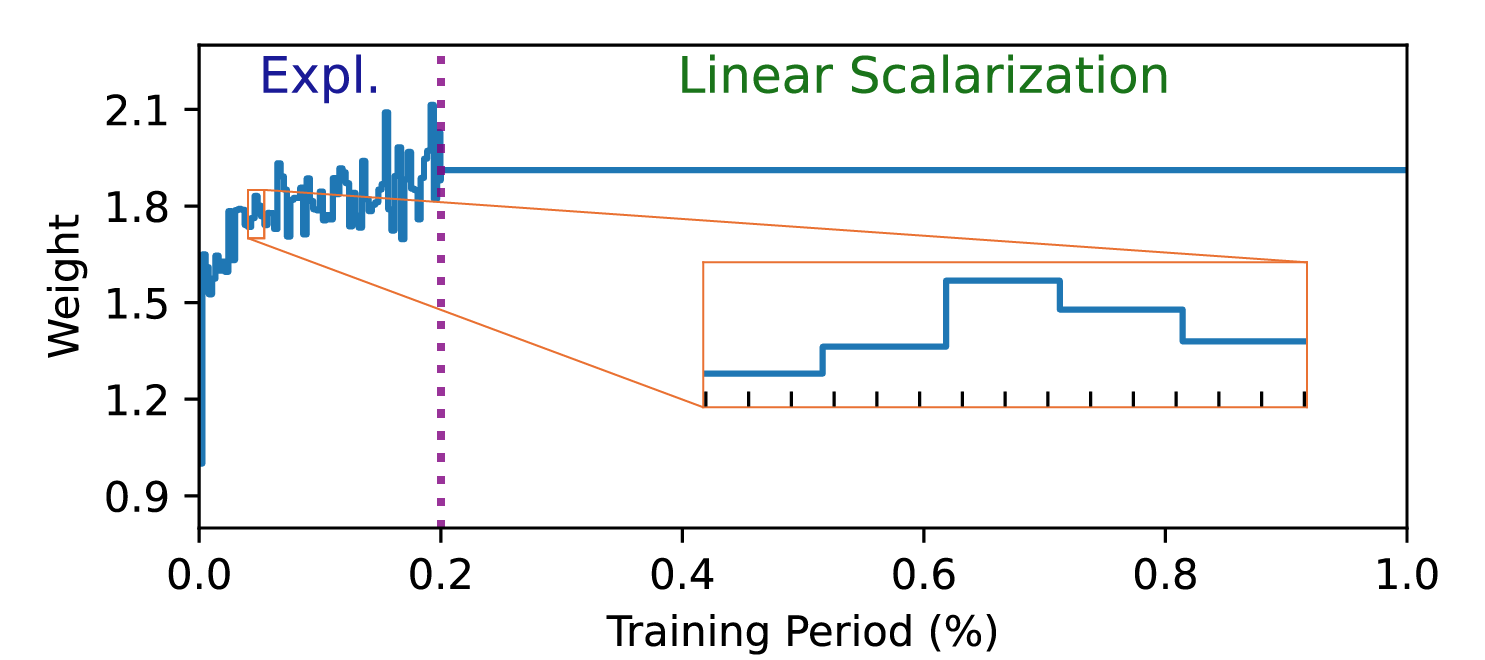}
    \caption{Weight evolution in AutoScale. AutoScale divides the training process into two phases: an exploration phase, where weights are iteratively computed in each local window, and a linear scalarization phase, where the final weight $\hat w$ is fixed
    for the remaining training.}
    \label{fig:method_illustration}
    \vspace{-15pt}
\end{figure}

Figure \ref{fig:method_illustration} shows how weight evolves in AutoScale. The magnified view highlights that the weight remains fixed within each local window during the exploration phase.

\begin{algorithm}[t]
\caption{AutoScale.}
\label{alg:cap}
\begin{algorithmic}
\Require Total iterations $T$, exploration ratio $\alpha$, window size $\tau$, cost function $\mathbf{F}(\mathrm{w})$, aggregation size $\eta$

\\

\State{\color{RoyalBlue} /* Phase 1: Exploration */}
\State{$\mathrm{w}^0 \gets \mathbf{1}$}
\State {\color{Blue} Partition into disjoint windows of size $\tau$}
\For {$i \gets 1:\alpha T / \tau$}

\State Gradient set $\mathcal{G}$ $\gets \emptyset$, Loss set $\mathcal{L} \gets \emptyset$

\For {$j \gets 1:\tau$}
\State{Run linear scalarization using $\mathrm{w}^{i-1}$}
\State $\mathcal{G} \gets \mathcal{G} \cup \{\mathbf{G}^j\}$, $\mathcal{L} \gets \mathcal{L} \cup \{\mathbf{L}^j\}$

\EndFor
\State {\color{Blue}Update weight for next window}
\State{$\mathrm{w}^i \leftarrow \argmin_\mathrm{w} \mathbf{F}(\mathrm{w}|\mathcal{G}, \mathcal{L}), \quad \textrm{s.t.} \;\,  \sum_{k=1}^K w_k = K$}

\EndFor

\\

\State{\color{RoyalBlue} /* Phase 2: Linear Scalarization  */}
\State{\color{Blue} Determine fixed weight for remaining $(1-\alpha)T$ iterations}
\State {$\mathrm{\hat w} \gets \frac{1}{\eta}\sum_{j=0}^{\eta-1} \mathrm{w}^{\alpha T / \tau-j}$}

\For {$t \gets \alpha T +1:T$}
\State {Run linear scalarization using $\hat{\mathrm{w}}$}
\EndFor
\end{algorithmic}
\label{algo:autoscale}
\end{algorithm}

\vspace{-20pt}

\begin{figure*}
    \centering
    \includegraphics[width=\linewidth]{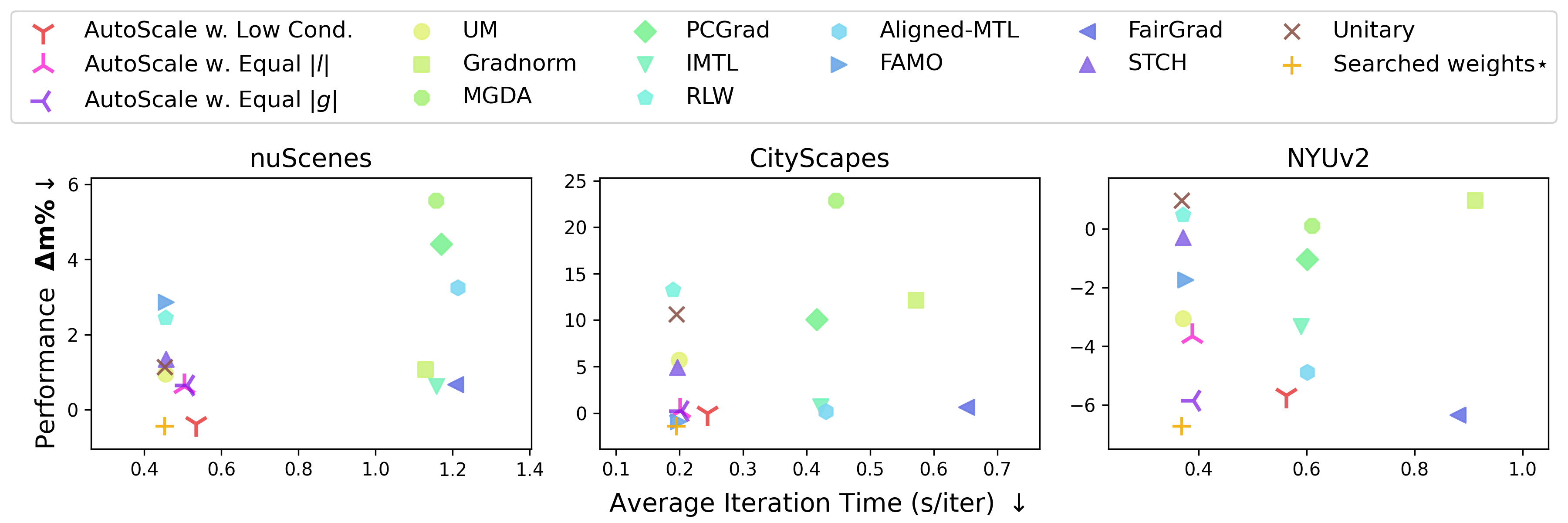} 
    \caption{
    Performance ($\mathbf{\Delta m\%}$) vs. training time.
    \scriptsize{$^\star$ The searched weights are from \textbf{20} trials, but only one trial's time is plotted; actual time is \textbf{20} times longer.}
    }
    \label{fig:performance_time}
\end{figure*}
\begin{table*}
\centering
\caption{\textbf{Perception of traffic sence}. \textsc{NuScenes}: two tasks, a large scale dataset, with Unitr~\citep{wang2023unitr} model.
\textsc{CityScapes}: three tasks with PSPNet~\citep{zhao2017pyramid} model.
Best scores are in \colorbox{mygray}{gray}, second-best in \textbf{bold}, and third-best \underline{underlined}.
\scriptsize{$^\star$The performance reported for the searched weights represents the best result from 20 search trials.$^\dagger$s/iter denotes seconds per training iteration; time required for the searched weights depends on the number of trails ($n$).}
}
\label{tab:res_nus_city}
\resizebox{0.93\linewidth}{!}{
\begin{tabular}
{l|ccc|ccc|c|ccc|ccc|c} 
\toprule
\multirow{3}{*}{Method} & \multicolumn{7}{c|}{\textbf{\textsc{NuScenes}}}                                                                                                                                                      & \multicolumn{7}{c}{\textbf{\textsc{CityScapes}}}                                                                                                                                                                                                         \\ 
\cmidrule(lr){2-8} 
\cmidrule(lr){9-15}
\multicolumn{1}{c|}{}                        & \multicolumn{2}{c}{3D Det. $\uparrow$} & Seg. $\uparrow$ & \textbf{MR}                                      & $\mathbf{\Delta m_{deg}}$                                & $\mathbf{\Delta m}$                                     & Time     & S. Seg.        & I. Seg.        & Disp.            & \textbf{MR}                          & $\mathbf{\Delta m_{deg}}$                                & $\mathbf{\Delta m}$                                     & Time    \\
\multicolumn{1}{c|}{}                        & mAP   & NDS                 & mIoU  & $\downarrow$                   & $\%\downarrow$                      & $\%\downarrow$                      & s/iter$^\dagger$ $\downarrow$   & mIoU $\uparrow$            & L1 $\downarrow$             & MSE $\downarrow$                 & $\downarrow$                    & $\%\downarrow$                      & $\%\downarrow$                       & s/iter$^\dagger$ $\downarrow$  \\ 
\midrule
STL Baseline                                 & 0.693 & 0.725               & 0.701 & -                                       & -                                          & -                                           & -        & 66.73                & 10.55                & 0.330                & -                                        &  -                                       & -                                           & -       \\ 
\midrule
\textit{\footnotesize{Linear Scalization}}   &       &                     &       &                                         &                                            &                                             &          &                      &                      &                      &                                          &                                            &                                             & \multicolumn{1}{l}{}        \\
Unitary                                      & 0.699 & 0.729               & 0.680 & 6.0                                     & 2.98                                     & 1.14                                      & 0.453    & 54.16                & 9.96                 & 0.392                & 9.00                                     & 18.74                                    & 10.62                                     & 0.195                       \\
Searched weights$^\star$                      & 0.695 & 0.725               & 0.706 & \textbf{3.0}                            & {\cellcolor{mygray}}0.00 & {\cellcolor{mygray}}-0.44 & 0.453 $\times n$ & 66.27                & 10.36                & 0.320                & {\cellcolor{mygray}}3.67 & {\cellcolor{mygray}}0.69 & {\cellcolor{mygray}}-1.42 & 0.195 $\times n$                \\ 
\midrule
\textit{\footnotesize{Various MTOs}}         &       &                     &       &                                         &                                            &                                             &          &                      &                      &                      &                                          &                                            &                                             & \multicolumn{1}{l}{}        \\
UM~\cite{Uncertainty}                 & 0.681 & 0.716               & 0.698 & 8.0                                     & 0.95                                     & 0.95                                      & 0.455    & 57.96                & 9.99                 & 0.361                & 8.00                                     & 11.20                                    & 5.69                                      & 0.199                       \\
Gradnorm~\cite{GradNorm}              & 0.677 & 0.714               & 0.700 & 7.0                                     & 1.07                                     & 1.07                                      & 1.130    & 52.53                & 10.06                & 0.395                & 11.00                                    & 20.49                                    & 12.11                                     & 0.572                       \\
MGDA~\citep{MGDA}                     & 0.647 & 0.696               & 0.660 & 14.0                                    & 5.57                                     & 5.57                                      & 1.157    & 67.29                & 17.77                & 0.333                & 9.00                                     & 34.74                                    & 22.88                                     & 0.446                       \\
PCGrad~\citep{PCGrad}                 & 0.671 & 0.711               & 0.657 & 13.5                                    & 4.41                                     & 4.41                                      & 1.170    & 54.52                & 10.04                & 0.385                & 9.00                                     & 17.51                                    & 10.07                                     & 0.416                       \\
IMTL-G~\citep{IMTL}                   & 0.690 & 0.720               & 0.696 & 8.0                                     & 0.63                                     & 0.63                                      & 1.158    & 65.44                & 10.70                & 0.326                & 9.67                                     & 1.68                                     & 0.71                                      & 0.422                       \\
RLW~\citep{PCGrad}                    & 0.699 & 0.723               & 0.664 & 8.0                                     & 5.27                                     & 2.45                                      & 0.455    & 52.69                & 10.12                & 0.405                & 11.33                                    & 21.92                                    & 13.27                                     & 0.190                       \\
Aligned-MTL~\citep{AlignedMTL}        & 0.664 & 0.706               & 0.680 & 12.5                                    & 3.25                                     & 3.25                                      & 1.213    & 66.05                & 10.69                & 0.324                & \uline{6.67}                             & 1.16                                     & 0.16                                      & 0.430                       \\
FAMO~\citep{FAMO}                     & 0.643 & 0.692               & 0.702 & 9.0                                     & 5.86                                     & 2.87                                      & 0.457    & 66.02                & 10.25                & 0.327                & 7.67                                     & \uline{1.07}                             & \textbf{-0.92}                            & 0.198                       \\
FairGrad~\citep{ban2024fair}          & 0.683 & 0.720               & 0.699 & 6.5                                     & 0.67                                     & 0.67                                      & 1.208    & 65.98                & 10.90                & 0.322                & 8.00                                     & 2.24                                     & 0.65                                      & 0.651                       \\
STCH~\citep{lin2024smooth}            & 0.676 & 0.711               & 0.697 & 10.0                                    & 1.35                                     & 1.35                                      & 0.457    & 68.06                & 12.49                & 0.325                & 7.00                                     & 18.42                                    & 4.92                                      & 0.197                       \\ 
\midrule
\textit{\footnotesize{AutoScale (Ours) w.}} &       &                     &       &                                         &                                            &                                             &          & \multicolumn{1}{l}{} & \multicolumn{1}{l}{} & \multicolumn{1}{l|}{} &                                          &                                            &                                             & \multicolumn{1}{l}{}        \\
Low Cond.                                    & 0.697 & 0.727               & 0.703 & {\cellcolor{mygray}}2.0 & {\cellcolor{mygray}}0.00 & \textbf{-0.37}                            & 0.535    & 66.07                & 10.62                & 0.32                 & \textbf{6.00}                            & \textbf{0.82}                            & \uline{-0.03}                             & 0.244                       \\
Equal $|l|$                  & 0.686 & 0.719               & 0.698 & 7.5                                     & 0.70                                     & \uline{0.63}                                & 0.504    & 66.04                & 10.73                & 0.32                 & 7.33                                     & 1.39                                     & 0.25                                      & 0.201                       \\
Equal $|g|$                  & 0.695 & 0.726               & 0.699 & \uline{5.0}                             & \uline{0.39}                             & 0.65                                      & 0.511    & 66.10                & 10.68                & 0.32                 & 6.67                                     & 1.07                                     & 0.20                                      & 0.203                       \\
\bottomrule
\end{tabular}
}
\end{table*}

\vspace{+20pt}
\subsection{Cost Functions}\label{sec:method:cost_F}

$\mathbf{F}(\mathrm{w})$ is highly adaptable and can be designed with different MTO metrics.
Here we focus on three aligned MTO metrics based on prior observations in \Cref{fig:cost_city}, and they measure distinct optimization challenges.
We define the corresponding cost functions below.
The cost function is computed as the average of per-iteration cost
\(\mathbf{F}^t(\mathrm{w})\) within a local window: $\mathbf{F}(\mathrm{w}) = \frac{1}{\tau}\sum_{t=i}^{i+\tau} \mathbf{F}^t(\mathrm{w}).$

\vspace{-11pt} 
\paragraph{Gradient Magnitude Similarity
(Equal $|g|$).
}
To maximize gradient magnitude similarity,
we design a straightforward cost function that penalizes larger variance in task gradients:
\begin{equation}
\begin{aligned}
 \mathbf{A}^t_{\text{row}(i,j), k} &= 
\begin{cases} 
|g^t_i| & \text{if } k = i \\
-|g^t_j| & \text{if } k = j \\
0 & \text{otherwise,}
\end{cases}
\\
\text{e.g. } ~
\mathbf{A}^t|_{K=3} &= \begin{bmatrix}
    |g^t_1| & -|g^t_2| & 0 \\ 
    |g^t_1| & 0 & -|g^t_3| \\ 
    0 & |g^t_2| & -|g^t_3|
\end{bmatrix},
\end{aligned}
\label{eq:F1}
\end{equation}
where $\mathbf{A}^t$, $\text{row}(i,j)$ refers to the row index assigned to the task pair $(i, j)$, $i \neq j$, and $g^t_i$ denotes gradient of task $i$ at iteration $t$.
According to this definition, the cost function is minimized when a set of task weights results in equal magnitudes for all re-scaled task gradients.

\vspace{-11pt} \paragraph{Loss Similarity (Equal $|l|$).}
The objective is to find a weight that optimally
balances the task loss scales.
It follows the same formulation as \Cref{eq:F1}, with gradient magnitudes $|g^t_k|$ replaced by loss scales $|l^t_k|$.

\vspace{-11pt} \paragraph{Condition Number Minimization (Low Cond.).}
We define $\mathbf{F}^t(\mathrm{w}) = \kappa(\mathbf{G}^t_\mathrm{w})$,
where $\kappa(\mathbf{X})$ denotes the condition number of a matrix $\mathbf{X}$, i.e. $\kappa(\mathbf{X}) =\frac{\sigma_{max}}{\sigma_{min}}$. $\mathbf{G}^t_\mathrm{w} = [w_1g^t_1 \; w_2g^t_2 \;... \;w_Kg^t_K]$ is the gradient matrix consisting of scaled task gradients. 
Unlike Align-MTL~\citep{AlignedMTL}, which manipulates both the direction and magnitude of the task gradients to minimize the condition number, we optimize by only rescaling the gradients using an appropriate weight set.

\begin{table*}
\centering
\caption{\textbf{Scene understanding} (\textsc{NYUv2}, three tasks). We report MTAN~\cite{liu2019end_mtan} model performance.
The best scores are provided in gray cell color;
the second best is \textbf{bolt}; the third best is \underline{underlined}.
\scriptsize{$^\star$The performance reported for the searched weights represents the best result from 20 search trials.$^\dagger$s/iter denotes seconds per training iteration; time required for the searched weights depends on the number of trails ($n$).}
}
\label{tab:nyuv2}
\resizebox{0.95\linewidth}{!}{
\begin{tabular}{l|ccccccccc|ccc|c} 
\toprule
\multicolumn{1}{c|}{\multirow{3}{*}{Method}} & \multicolumn{2}{c}{Segmentation}                                              & \multicolumn{2}{c}{Depth}                                                  & \multicolumn{5}{c|}{Surface Normal}                                     & \multicolumn{1}{l}{}    & \multicolumn{1}{l}{}      & \multicolumn{1}{l}{}     & \multicolumn{1}{l}{}                  \\ 
    \cmidrule(lr){2-3}\cmidrule(lr){4-5}\cmidrule(lr){6-10}
\multicolumn{1}{c|}{}                        & \multicolumn{1}{l}{\multirow{2}{*}{mIoU~$\uparrow$}} & \multicolumn{1}{l}{\multirow{2}{*}{Pix Acc~$\uparrow$}} & \multicolumn{1}{l}{\multirow{2}{*}{Abs.~$\uparrow$}} & \multicolumn{1}{l}{\multirow{2}{*}{Rel.~$\uparrow$}} & \multicolumn{2}{c}{Angle Dist.~$\downarrow$} & \multicolumn{3}{c|}{within $t^\circ$~$\uparrow$} & \textbf{MR}             & $\mathbf{\Delta m_{deg}}$ & $\mathbf{\Delta m}$      & Time                                  \\
\multicolumn{1}{c|}{}                        & \multicolumn{1}{l}{}                      & \multicolumn{1}{l}{}                         & \multicolumn{1}{l}{}                      & \multicolumn{1}{l}{}                      & Mean  & Median                  & 11.25 & 22.5 & 30                     & $\downarrow$            & $\%\downarrow$              & $\%\downarrow$             & s/iter$^\dagger$ $\downarrow$         \\ 
\midrule
STL Baseline                                 & 40.14 & 65.43                                     & 0.61  & 0.26                               & 24.10 & 17.47                   & 0.33  & 0.60 & 0.72           & -                                        & -                                        & -                                         & \multicolumn{1}{l}{-}          \\ 
\midrule
\textit{\footnotesize{Linear Scalization}}   &       &                                           &       &                                    &       &                         &       &      &                & \multicolumn{1}{l}{}                     & \multicolumn{1}{l}{}                     & \multicolumn{1}{l}{}                      & \multicolumn{1}{l}{}           \\
Unitary                                      & 40.89 & 67.10                                     & 0.52  & 0.23                               & 27.26 & 22.39                   & 0.25  & 0.50 & 0.63           & 11.67                                    & 18.88                                    & 0.97                                      & 0.369                          \\
Searched weights$^\star$                     & 44.25 & 68.60                                     & 0.52  & 0.22                               & 24.42 & 18.41                   & 0.31  & 0.58 & 0.70           & \uline{4.33}                             & \textbf{3.29}                            & {\cellcolor{mygray}}-6.73 & \multicolumn{1}{l}{0.369 $\times n$}  \\ 
\midrule
\textit{\footnotesize{Various MTOs}}         &       &                                           &       &                                    &       &                         &       &      &                &                                          &                                          &                                           &                                \\
UM~\cite{Uncertainty}                                           & 41.74 & 67.63                                     & 0.51  & 0.22                               & 26.07 & 20.56                   & 0.27  & 0.54 & 0.67           & 6.67                                     & 12.06                                    & -3.06                                     & 0.371                          \\
Gradnorm~\cite{GradNorm}                                     & 40.77 & 66.96                                     & 0.52  & 0.22                               & 27.63 & 22.88                   & 0.24  & 0.49 & 0.62           & 11.00                                    & 20.91                                    & 0.96                                      & 0.912                          \\
MGDA~\citep{MGDA}                                         & 32.20 & 62.48                                     & 0.55  & 0.23                               & 23.97 & 17.64                   & 0.33  & 0.60 & 0.71           & 10.33                                    & 6.24                                     & 0.11                                      & 0.610                          \\
PCGrad~\citep{PCGrad}                                       & 42.03 & 67.95                                     & 0.54  & 0.23                               & 26.48 & 21.23                   & 0.26  & 0.53 & 0.65           & 9.00                                     & 14.53                                    & -1.04                                     & 0.601                          \\
IMTL-G~\citep{IMTL}                                       & 41.36 & 67.09                                     & 0.52  & 0.22                               & 25.35 & 19.72                   & 0.29  & 0.56 & 0.68           & 8.33                                     & 8.70                                     & -3.32                                     & 0.590                          \\
RLW~\citep{RLW}                                          & 41.80 & 67.07                                     & 0.53  & 0.22                               & 27.29 & 22.39                   & 0.25  & 0.50 & 0.63           & 10.67                                    & 18.95                                    & 0.48                                      & 0.371                          \\
Aligned-MTL~\citep{AlignedMTL}                                  & 40.55 & 67.09                                     & 0.50  & 0.21                               & 24.86 & 19.09                   & 0.30  & 0.57 & 0.69           & 6.67                                     & 6.00                                     & -4.88                                     & 0.601                          \\
FAMO~\citep{FAMO}                                         & 41.44 & 67.35                                     & 0.52  & 0.21                               & 26.60 & 21.56                   & 0.26  & 0.52 & 0.65           & 8.33                                     & 15.81                                    & -1.73                                     & 0.376                          \\
FairGrad~\citep{ban2024fair}                                     & 40.33 & 67.04                                     & 0.51  & 0.21                               & 24.04 & 17.84                   & 0.32  & 0.60 & 0.71           & 6.67                                     & {\cellcolor{mygray}}1.04 & \textbf{-6.35}                            & 0.880                          \\
STCH~\citep{lin2024smooth}                                         & 39.92 & 66.34                                     & 0.53  & 0.22                               & 26.36 & 21.33                   & 0.26  & 0.52 & 0.65           & 12.00                                    & 14.90                                    & -0.29                                     & 0.371                          \\ 
\midrule
\textit{\footnotesize{Auto Scale (Ours) w.}} &       &                                           &       &                                    &       &                         &       &      &                &                                          &                                          &                                           &                                \\
Low Cond.                                    & 42.32 & 68.03                                     & 0.50  & 0.21                               & 25.12 & 19.23                   & 0.30  & 0.57 & 0.69           & {\cellcolor{mygray}}3.33 & 6.79                                     & -5.67                                     & 0.562                          \\
Equal $|l|$                  & 40.83 & 66.81                                     & 0.50  & 0.21                               & 25.44 & 20.04                   & 0.28  & 0.55 & 0.68           & 7.67                                     & 9.81                                     & -3.65                                     & 0.389                          \\
Equal $|g|$                  & 41.65 & 67.74                                     & 0.50  & 0.21                               & 24.89 & 18.87                   & 0.30  & 0.57 & 0.69           & {\cellcolor{mygray}}3.33 & \uline{5.39}                             & \uline{-5.85}                             & 0.391                          \\
\bottomrule
\end{tabular}
}
\end{table*}
\section{Experiments}
We demonstrate the effectiveness of AutoScale by comparing it with various baseline methods on different benchmark datasets.

\vspace{-11pt} \paragraph{Datasets and Models.}
AutoScale is evaluated on three datasets: CityScapes~\citep{cordts2016cityscapes}, NYUv2~\citep{silberman2012indoor},  and NuScenes~\citep{caesar2020nuscenes}.

CityScapes and NYUv2 are wildly used MTL benchmarks, containing street-view and indoor RGB-D images with per-pixel annotations, respectively.
Following~\citet{AlignedMTL}, we adopt the same evaluation settings for both datasets, with three tasks (disparity estimation, instance segmentation, and semantic segmentation) for CityScapes and three tasks (depth estimation, semantic segmentation, and surface normal estimation) for NYUv2. 

In addition to standard benchmarks, we evaluate NuScenes, a large-scale autonomous driving dataset with over 40k annotated frames, each containing six camera images and a 32-beam LiDAR point cloud.
To the best of our knowledge, this is the first systematic benchmarking of MTOs at this scale in autonomous driving.
Compared to CityScapes and NYUv2, NuScenes has higher task complexity and requires 156 training hours per run, which is $\sim$\textbf{30}$ \times$ more than the training time of other two datasets. We focus on two tasks: 3D object detection and bird-eye-view (BEV) map segmentation.

For model architectures, we use PSPNet~\citep{zhao2017pyramid} for CityScapes,  MTAN~\citep{liu2019end_mtan} for NYUv2,  UniTR~\citep{wang2023unitr} for NuScenes.

Further details of the experiment setup can be found in supplementary.

\vspace{-11pt} \paragraph{Baseline Methods.}
We compare AutoScale algorithm with 13 other methods: single-task learning (STL), 
UM~\citep{Uncertainty}, GradNorm~\citep{GradNorm}, MDGA~\citep{MGDA}, IMTL-G~\citep{IMTL}, PCGrad~\citep{PCGrad}, RLW~\citep{RLW}, Aligned-MTL~\citep{AlignedMTL}, FAMO~\citep{FAMO}, STCH~\citep{lin2024smooth}, FairGrad~\citep{ban2024fair}, 
linear scalarization with unitary weighting ($w_k = 1,\; \forall 1\leq k \leq K$), and linear scalarization with the best-performing weight found via grid search with 20 trails.

\vspace{-11pt} \paragraph{Evaluation Metrics.} \label{sec:eval_metrics}Following previous methods~\citep{AlignedMTL, FAMO}, we use $\mathbf{\Delta m}$ and Mean Rank (MR) metrics to evaluate multi-task performance.
\textbf{1)} $\mathbf{\Delta m}$ measures the average performance drop relative to the single-task baseline across all tasks and is defined as
$\mathbf{\Delta m}= \frac{1}{K} \sum^{\scaleto{K\mathstrut}{5pt}}_{k=1}(-1)^{\sigma_k}\mathbf{\delta m_k}$. Here, we denote $\mathbf{\delta m_k}=\frac{M_k-B_k}{B_k} \times 100$ as the performance difference on task $k$, where $M_k$ and $B_k$ are the $k$-th task metric evaluated on a multi-task model and a single-task baseline respectively. $\sigma_k = 1$ if higher value of $M_k$ is better, and $\sigma_k = 0$ otherwise. \textbf{2) Mean Rank (MR)} is the average ranking of performance across all tasks over all methods. For example, if a method ranks first on one task but second on the other task, $\text{MR}=(1+2)/2=1.5$.

While $\mathbf{\Delta m}$ captures the average performance change, it fails to distinguish between improvements and degradations.
To explicitly quantify total performance drop, we introduce, 
a new metric $\mathbf{\Delta m_{deg}}$, which sums up all per-task performance decreasing $\delta \mathbf{m}$, that is, $\mathbf{\Delta m_{deg}}=\sum_{{\scaleto{k=1\mathstrut}{6pt}}}^{{\scaleto{K\mathstrut}{4pt}}}\max((-1)^{\sigma_k}\mathbf{\delta m_k}, 0)$.
This metric captures the total percentage of performance drops ($(-1)^{\sigma_k}\mathbf{\delta m_k} > 0$) while ignoring improvements ($(-1)^{\sigma_k}\mathbf{\delta m_k} < 0$).
By focusing only on degradations, it is useful in cases where minimizing overall performance drops is prioritized over sacrificing some tasks' performance to enhance others.

\vspace{-11pt} \paragraph{Default Settings.} For all experiments on three benchmark datasets shown in \Cref{tab:res_nus_city} and \Cref{tab:nyuv2}, we use the following parameter settings for AutoScale: the exploration ratio $\alpha=0.2$; window size $\tau=50$; and aggregation size $\eta=10$.
The cost function is minimized using Sequential Least Squares Programming (SLSQP)~\cite{kraft1988software}.
\label{subsec:result}

\vspace{-11pt} \paragraph{Results and Efficiency.}
\Cref{tab:res_nus_city} and \Cref{tab:nyuv2} present the quantitative comparison of AutoScale against state-of-the-art methods on the NuScenes, CityScapes, and NYUv2 datasets, while \Cref{fig:performance_time} summarizes performance ($\mathbf{\Delta m\%}$) versus training efficiency.
Across all datasets, AutoScale consistently ranks among the top-performing methods in terms of MR, $\mathbf{\Delta m_{deg}\%}$, $\mathbf{\Delta m\%}$, and time efficiency. 

On NuScenes and CityScapes, AutoScale, using the low condition number metric as an indicator, achieves superior results compared to all other MTO methods.
It even surpasses searched weights in MR and $\mathbf{\Delta m_{deg}\%}$ on large-scale NuScenes dataset. 
For NYUv2, it remains highly competitive, outperforming most MTO methods and achieving results comparable to FairGrad.

Regarding efficiency, AutoScale benefits from its second-phase linear scalarization design.
It is 2–3× faster than gradient-manipulating MTOs including FairGrad, Aligned-MTL, ITML, GradNorm, MGDA, and PCGrad.

In summary, AutoScale achieves the best trade-off between performance and efficiency compared to SOTA.
Efficiency-focused methods like FAMO and STCH are fast but less robust across datasets.
While grid-searched weights achieve the best performance, they require extensive tuning. 
AutoScale demonstrates the effectiveness of using MTO metrics for automatic weight selection and achieves performance closest to grid-searched scalarization.

\begin{figure}[t]
\centering

\begin{subfigure}[t]{1\linewidth}
\centering
    {\includegraphics[width=0.93\linewidth]{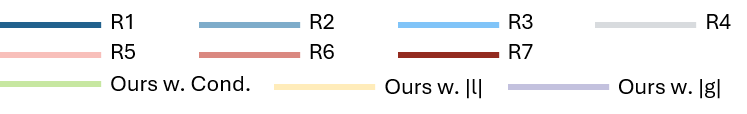}}
\end{subfigure}
\hfill
\begin{subfigure}[t]{.48\linewidth}
\centering
    \includegraphics[width=1.0\linewidth]{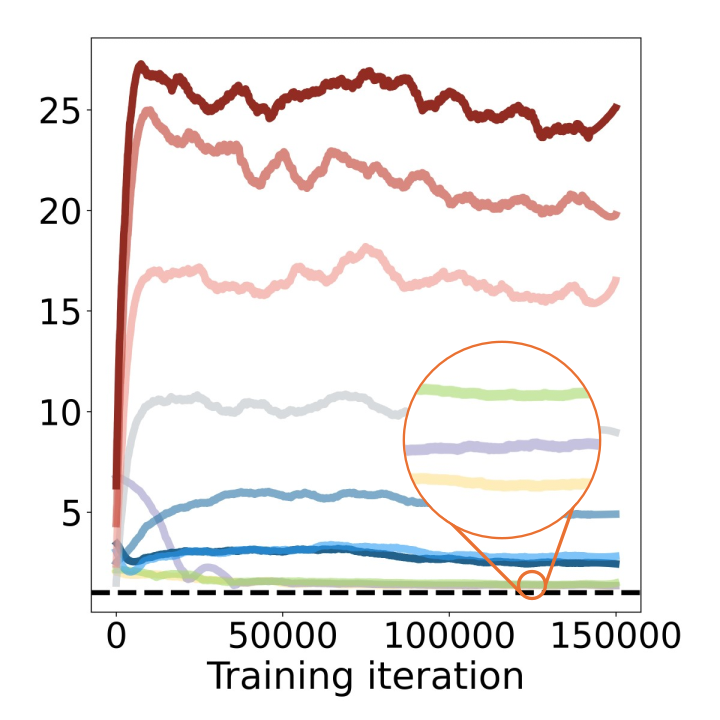}
    \caption{Condition Number}
    \label{fig:supply:city:2:cnumber}
\end{subfigure}
\begin{subfigure}[t]{.48\linewidth}
    \centering
    \includegraphics[width=1.0\linewidth]{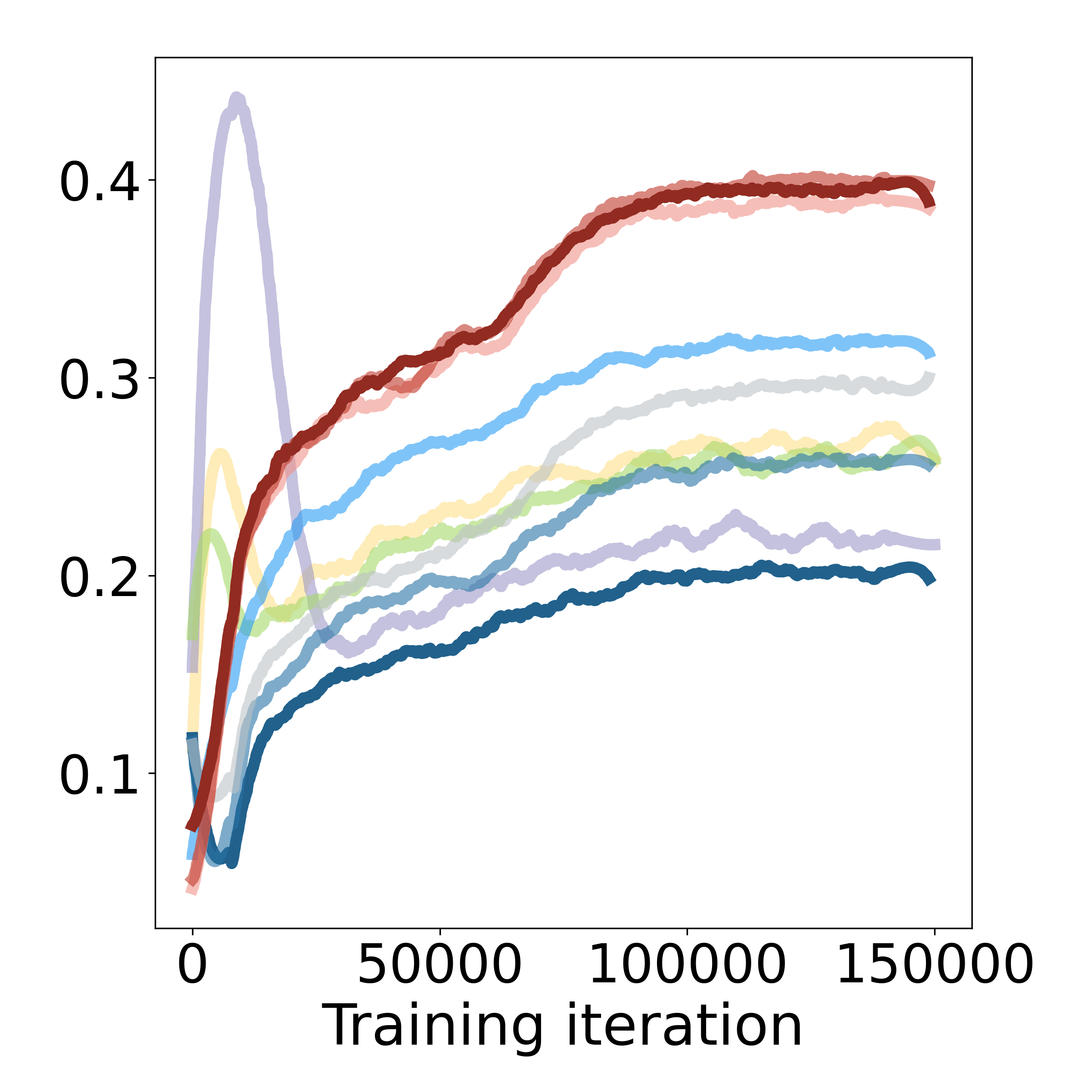}
    \caption{Std of inverse learning rate}
    \label{fig:supply:city:3:ilr}
\end{subfigure}
\caption{How metrics evolve during AutoScale training? In addition to seven linear scalarization weight sets on the CityScapes dataset in \Cref{fig:eval_city}
, we include our AutoScale to observe how metrics behave.
The performance ranking of all runs based on $\mathbf{\Delta m}$ is: R1 $>$ R2 $>$ Ours w. Cond. $>$ Ours w. $|g|$ $>$ Ours w. $|l|$ $>$R3 $>$ R4 $>$ R5 $>$ R6 $>$ R7.
AutoScale exhibits favorable trends across different metrics.}
\label{app:fig:eval_city_w_ours}
\vspace{-16pt}
\end{figure}

\vspace{-11pt} \paragraph{Ablation Study.}
We conduct ablation studies on the exploration ratio $\alpha$ and 
aggregation size $\eta$
of AutoScale, using default settings outlined in \Cref{subsec:result} with MTAN~\citep{liu2019end_mtan} on NYUv2 dataset~\citep{silberman2012indoor}. Detailed results are provided in the supplementary.

\vspace{-11pt} \paragraph{How metrics evolve during AutoScale training?}
We further analyze how MTO metrics evolve 
during AutoScale training with three cost functions, as shown in Figure \ref{app:fig:eval_city_w_ours} (and supplementary).
AutoScale consistently exhibits favorable trends across different metrics, including high gradient similarity, low condition number, balanced convergence speed (inverse learning rate), balanced loss scale, and equal angles of the final aggregated gradient.
These trends hold regardless of whether the cost function optimizes equal gradient scale, equal loss scale, or low condition number.
These results further validate our hypothesis that MTO metrics play a crucial role in finding well-performing scalarization weights. 

When comparing the three cost functions, we observe that optimizing loss scale introduces stronger fluctuations in the condition number and inverse learning rate during the early stage of training, compared to the other two MTO metrics.
Given the observed performance ranking of the three cost functions i.e. Low Cond. $>$ Equal $|g|$ $>$ Equal $|l|$, it is reasonable to infer that these fluctuations contribute to the lower performance of Equal $|l|$ relative to the other two cost functions.

These findings indicate that the values of MTO metrics are interrelated, influencing each other, and a well-performing training run requires alignment across multiple MTO metrics. This interdependence will potentially be the focus of future works in this area of MTL. 

\vspace{-9pt}

\paragraph{Comparison with Weight Search Methods.} We additionally evaluate our methods against two performance-base weight search methods - Bayesian optimization and population-based training in supplementary. Through this, we highlight the efficiency of our weight selection mechanism.

\vspace{-7pt}

\section{Conclusion}
\vspace{-5pt}
In this work, we propose a novel perspective for understanding linear scalarization via MTO metrics. 
Through comprehensive experiments, we identify that well-performing linear scalarization aligns with specific characteristics of certain MTO metrics, including high gradient magnitude similarity, low condition number, and more balanced loss scale across tasks.
Building on these insights, we introduce AutoScale, an efficient framework that automatically determines optimal weights by optimizing MTL metrics in two phases within a single training run. Our experiments demonstrate that
AutoScale achieves superior performance and high efficiency across a wide range of benchmarks including a large-scale modern autonomous driving dataset, closest to the searched weight, but without the need of grid search. 

\paragraph{Acknowledgements} The computations were enabled by the supercomputing resource Berzelius provided by National Supercomputer Centre at Linköping University and the Knut and Alice Wallenberg foundation, Sweden.

{
    \small
    \bibliographystyle{ieeenat_fullname}
    \bibliography{main}
}

\end{document}